\documentclass[journal,10pt]{IEEEtran}
\usepackage[utf8]{inputenc} 
\usepackage[T1]{fontenc}    
\usepackage{hyperref}       
\usepackage{url}            
\usepackage{booktabs}       
\usepackage{amsfonts}       
\usepackage{nicefrac}       
\usepackage{microtype}      

\usepackage[fleqn]{amsmath}
\usepackage{amsfonts}
\usepackage{amssymb}
\usepackage{bbm}  

\usepackage{graphicx}
\usepackage{amsthm}
\usepackage[noadjust]{cite}

\newtheorem{theorem}{Theorem}

\newtheorem{definition}[theorem]{Definition}
\newtheorem{proposition}[theorem]{Proposition}
\newtheorem{lemma}[theorem]{Lemma}

\usepackage{array}

\usepackage[caption=false,font=footnotesize]{subfig}

\usepackage{url}

\usepackage{float}
\newfloat{footnote}{hb}

\DeclareMathOperator*{\argmin}{arg\,min}

\newcommand{\st}{\;\text{ s.t. }}
\newcommand{\diag}{\text{diag}}

\newcommand{\SNR }{\text{SNR}}

\newcommand{\Null}{\textup{Null}}

\def \zero {{\mathbf 0 }}
\def \Const {{\beta }}

\def \transpose {{\boldsymbol T }}

\def \' {^{*}}

\def \A {{\boldsymbol A}}

\def \AJ {{\boldsymbol A^{\boldsymbol J } }}

\def \a {{\boldsymbol a}}

\def \B {{\mathcal B}}

\def \cOne    {{\mathcal{C}_0 }}
\def \cTwo    {{\mathcal{C}_1 }}

\def \D  {{\boldsymbol D}}

\def \E {{\mathbb E}}  
\def \e {{\boldsymbol e}}

\def \Identity {{\boldsymbol  I}}
\def \I {{\mathcal I}}

\def \M  {{\boldsymbol M}}
\def \m  {{\boldsymbol m}}

\def \O {\mathcal{O}}

\def \P {{\mathbb P}}  

\def \Q {{\boldsymbol Q}}
\def \QJ {{\boldsymbol Q^{\boldsymbol J } }}

\def \R {\mathbb{R}}

\def \S {\mathcal{S}}

\def \sparse {s}

\def \v  {{\boldsymbol v}}
\def \V  {{\boldsymbol V}}

\def \X  {{\boldsymbol X}}
\def \optX { \boldsymbol{X^{\star}}}

\def\x  {{\boldsymbol x}}
\def \optx { \boldsymbol{x^{\star}}}
\def\xf  {{\boldsymbol {x^J}}}
\def\xB  {{\boldsymbol {x^B}}}

\def\xci  {{ \x_{\ci} }}

\def \xtmp  {{\boldsymbol{x_0}}} 

\def \Xtmp { {\boldsymbol{\X_0}}}  
\def \Xsharp { \X^{\boldsymbol \sharp} }

\def \Xwhatever            { \X^{\boldsymbol{ J \star}} }

\def \y {{\boldsymbol y}}

\def \z {{\boldsymbol z}}

\def \Z {{\boldsymbol Z}}

\def \zero {{\mathbf 0 }}
\def \Const {{\beta }}

\def \T {{\mathcal S } }

\def \diffX {\Delta}

\def \lonemin {\mathrm{ \mathbf{P_1^{\boldsymbol{\epsilon}}}}}

\def \ours {\mathrm{ \mathbf{J^{\boldsymbol { \epsilon^{J} }  }_{12}} }}
\def \oursNoNoise {\mathrm{ \mathbf{J_{12}}}}

\def \loneminNoNoise {\mathrm{ \mathbf{P_1}}}

\def \loneminNoNoiseMulti  {\mathrm{ \mathbf{P_1}}(K)}
\def \lTwoOneNoNoise {\mathrm{ \mathbf{P_{12}}}(K)}

\def \indicator { \mathbbm{1} }

\def \ri  {i}
\def \ci  {j}

\def \xri                 {{ \x \text{\footnotesize{$[\ri]$}} }}
\def \xrOne               {{ \x \text{\footnotesize{$[1]$}} }}
\def \xrTwo               {{ \x \text{\footnotesize{$[2]$}} }}
\def \xrn               {{ \x \text{\footnotesize{$[n]$}} }}
\def \xci                {{ \x_j }}
\def \yIci               {{ \y \text{\scriptsize $[\I_\ci]$} }}
\def \AIci              {{ \A{ [\I_{\ci}]}  }}
\def \zIci               {{ \z{ \text{\scriptsize $[\I_{\ci}]$ } } }}

\def \zrOne                 {{ z \text{\footnotesize{$[1]$}} }}
\def \zrTwo                 {{ z \text{\footnotesize{$[2]$}} }}
\def \zrm                 {{ z \text{\footnotesize{$[m]$}} }}

\def \AEi              {{ \A{ [\I]}  }}
\def \ari               {{ \a \text{\footnotesize{$[\ri]$}} }}
\def \yEi               {{ \y{ \text{\scriptsize $[\I]$ } } }}
\def\xEi                 {{ \x \text{\scriptsize $[\I]$} }}
\def\zEi                 {{ \z \text{\scriptsize $[\I]$} }}
\def\xSolEi                 {{ \x \text{\scriptsize $(\I)$} }}

\def \yIcOne               {{ \y \text{\scriptsize $[\I_1]$} }}
\def \AIcOne              {{ \A{ [\I_{1}]}  }}
\def \zIcOne               {{ \z \text{\scriptsize $[\I_1]$} }}

\def \yIcTwo               {{ \y \text{\scriptsize $[\I_2]$} }}
\def \AIcTwo              {{ \A{ [\I_{2}]}  }}
\def \zIcTwo               {{ \z \text{\scriptsize $[\I_2]$} }}

\def \yIcK               {{ \y \text{\scriptsize $[\I_K]$} }}
\def \AIcK              {{ \A{ [\I_{K}]}  }}
\def \zIcK               {{ \z \text{\scriptsize $[\I_K]$} }}

\title{JOBS: Joint-Sparse Optimization from Bootstrap Samples}
\author{Luoluo~Liu,~\IEEEmembership{Student Member,~IEEE,}
        Sang~(Peter)~Chin,~\IEEEmembership{Member,~IEEE,}
        Trac~D.~Tran,~\IEEEmembership{Fellow,~IEEE}%
\thanks{Luoluo Liu is with Department of Electrical Engineering, Johns Hopkins University, Baltimore, MD, 21210}
\thanks{Prof. S. Chin is with Department of Computer Science $\&$ Hariri Institute of Computing, Boston University, Boston, MA, 02215 and Department of Electrical Engineering, Johns Hopkins University, Baltimore, MD, 21210}
\thanks{Prof. T. Tran is with Department of Electrical Engineering, Johns Hopkins University, Baltimore, MD, 21210}
\thanks{This work is partially supported by National Science Foundation
under Grants xxx and Air Force Office of Scientific Research xxx, Manuscript revised Nov 16, 2018
}}
\begin{document}
%
\pdfoutput=1
\maketitle
\addtolength{\belowcaptionskip}{-5mm}
\begin{abstract}
Classical signal recovery based on $\ell_1$ minimization solves the least squares problem with all available measurements via sparsity-promoting regularization. In practice, it is often the case that not all measurements are available or required for recovery. Measurements might be corrupted/missing or they arrive sequentially in streaming fashion.
In this paper, we propose a global sparse recovery strategy based on subsets of measurements, named JOBS, in which multiple measurements vectors are generated from the original pool of measurements via bootstrapping, and then a joint-sparse constraint is enforced to ensure support consistency among multiple predictors. The final estimate is obtained by averaging over the $K$ predictors.
The performance limits associated with different choices of number of  bootstrap samples $L$ and number of estimates $K$ is analyzed theoretically. Simulation results validate some of the theoretical analysis, and show that the proposed method yields state-of-the-art recovery performance, outperforming $\ell_1$ minimization and 
other existing bootstrap-based techniques in the challenging case of low levels of measurements. Our proposed framework is also preferable over other bagging-based methods in the streaming setting since it yields better recovery performances
with small $K$ and $L$ for data-sets with large sizes.
\end{abstract}

\section{Introduction}
\label{sec:intro}
In Compressed Sensing (CS) and sparse recovery, solutions to the linear inverse problem in the form of least squares plus a sparsity-promoting penalty term have been intensively studied.
Formally speaking, a the measurements vector $\y \in \R^m $ is generated by $\y = \A \x + \z$, where $\A \in \R^{m \times n}$ is the sensing matrix, $\x \in \R^n$ is the sparse coefficient with very few non-zero entries and $\z$ is a bounded noise vector.
The problem of interest is finding the sparse vector $\x$ given $\A$ as well as  $\y$. However, directly minimizing the support size is proven to be NP-hard~\cite{sasNPhard}. Instead, a convex regularizer is preferable. Among various choices, the $\ell_1$ norm 
 is the most commonly used. The noiseless case is referred to as \emph{Basis Pursuit} (BP):
\begin{equation}
\label{eq:smv_eq}
\loneminNoNoise:  \min  \| \x \|_1  \st  \y = \A \x.
\end{equation}
The noisy version is known as \emph{basis pursuit denoising}  \cite{bpdn},
or \emph{least absolute shrinkage and selection operator} (LASSO)
\cite{lasso}:
\begin{equation}
\label{eq:smv_ineq}
\lonemin: \min  \| \x \|_1  \st \| \y - \A \x \|_2 \leq \epsilon.
\end{equation}

The performance of $\ell_1$ minimization in recovering the true sparse solution 
has been thoroughly investigated
 in CS literature~\cite{cs,CandesRobustUP,donohoCS,CSincoherence07}.  
CS theory reveals that when the true solution is sparse and if the number of measurements is large enough, then the solution to (\ref{eq:smv_eq}) converges to the ground truth and (\ref{eq:smv_ineq}) converges to its neighbourhood with high probability~\cite{cs}.

Unfortunately, in practice, all measurements may not be available. Some parts of the data can be missing or severely corrupted. 
In streaming settings, measurements might be available sequentially or in small batches. Wasting valuable time and buffering memory might not be the optimal strategy.

Alternatively, for sparse recovery or sparse-representation-based classification, many schemes use local observations and show promising performances~\cite{ksvd,super_resolution,pfr,yiface}.
It is not surprising since the number of measurements collected is usually much larger than lower bounds suggested by theory.
However, proper choices of subset(s) differ between applications and require case-by-case treatment.
Prior knowledge helps significantly in the selection process.
For example, image data-sets may have large variance overall but relative invariance within local regions, choosing to work with image patches performs well in dictionary learning and deep learning~\cite{ksvd,alexnet}.

Without any prior information, a natural method is sampling uniformly at random with replacement, termed \emph{bootstrap}~\cite{efron_bootstrap}.
It performs reasonably well when all measurements are equally good. 
In CS theory, 
some random matrices have been proven to be good sensing matrices.
These operators act by shuffling and recombining entries of the original measurements. Consequently, any spatial or temporal structure 
would be destroyed, making the measurements even more democratic.

To incorporate the information from multiple estimates, 
the \emph{Bagging}~\cite{bagging} procedure was proposed. It solves 
objectives multiple times \emph{independently} from bootstrap samples and then averages over multiple predictions.
Applying Bagging in sparse regression 
was shown to reduce estimation error when the sparsity level $\sparse$ is high~\cite{bagging}. However, individually solved predictors aren't guaranteed to have the same support and in the worst case, their average can be quite dense: with its support size growing up to $K \sparse$. To alleviate this problem, Bolasso was proposed~\cite{bolasso},
which firstly recovers the support of the final estimate by detecting the common support among $K$ individually solved predictors generated from bootstrap and then applies least squares on the common support. However, this strategy is very aggressive. When the noise level is high, it commonly recovers the zero solution.

In this paper, to resolve the support consistency issue in previous approaches and avoid issues caused by a two-step process, we propose to enforce
row sparsity among all predictors using the $\ell_{1,2}$ norm within the iterative optimization loop.
The entire process is as follows. 
First, we draw $L$ samples from $m$ measurements with replacement. 
Then we repeat this sampling process $K$ times to generate 
$\I_{1}, \I_{2}, ..., \I_{K}$, each of size $L$. This sampling process returns $K$ 
 multi-sets of the original data $\{ \yIcOne , \AIcOne \} , \{\yIcTwo, \AIcTwo \} ...., \{ \yIcK, \AIcK \} $. Here we introduce the notation $ (\boldsymbol{\cdot})[{\I}]$ : for a set (multi-set) $\I$, the operation $ [{\I}]$ takes \emph{rows} supported on $\I$ and throws away all other rows in the complement $\I^c$. For each solution $\xci$ that corresponds to its data pair $\{ \yIci , \AIci \} $, we enforce the row sparsity constraint $\ell_{1,2}$ penalty on them to enforce the same support among all predictors.
The $\ell_{1,2}$ norm penalty is defined as: $\|\X\|_{1,2} = \sum_{\ri} \| \xri^{\transpose} \|_2 $, where $\xri$ denotes the $\ri-$th row of $\X$. The final estimates $\x^{\boldsymbol{J}}$ is obtained by averaging
over all $K$ estimators. 
We coin the whole procedure JOBS (\textbf{J}oint-sparse \textbf{O}ptimization from \textbf{B}ootstrap \textbf{S}amples). Other choices of row sparsity convex norms are suggested 
in~\cite{chmmv, blocksparse,tropp_oneinf, Mfocuss}. 

The main contributions of this paper are:
{\it (i)} We propose and demonstrate JOBS, employing the powerful bootstrapping, inspired from machine learning,
and improves robustness of sparse recovery in noisy environments through the use of a collaborative recovery scheme.
{\it (ii) } We explore the proposed strategy in-depth. Since the key parameters in our method is the bootstrap sample size $L$ and the number of bootstrap samples $K$, we derive various error bounds analytically with regards to these parameters. 
{\it (iii)} We also study optimal parameter settings and validate theory via extensive simulations.

For fair comparison to our method, we also extend and study 
Bagging and Bolasso, in the same setting.
Solutions $ \x^{\boldsymbol{\sharp}}_1, \x^{\boldsymbol{\sharp}}_2, ... \x^{\boldsymbol{\sharp}}_K $ solved independently from the same observation as JOBS: 
$\{ \yIcOne , \AIcOne \} , \{\yIcTwo, \AIcTwo \} ...., \{ \yIcK, \AIcK \} $. Bagging takes average of multiple estimates and Bolasso conducts post-processing to ensure the support consistency of solution.
Further contributions are:
{\it (iv)}  We explore the theoretical analysis for employing Bagging in sparse recovery.
{\it (v)} Although the original Bagging and Bolasso use bootstrap ratio $L/m = 1$, we studied the behavior of these two algorithms with multiple ratios $L/m$ from $0$ to $1$, same as JOBS, to explore the optimal parameters as well as to make a fair comparison.
{\it (vi)} We study a subsampling variation of the proposed scheme as an alternative to bootstrapping by simulations.

Simulation results show that our methods outperform all other methods when the number of measurements is small. While the number of measurements is large, acceptable performance of JOBS can be obtained with very small $L$ and $K$ and outperform Bagging and Bolasso,
which potentially has an advantage in streaming settings in which $\ell_1$ minimization is not applicable and JOBS can achieve acceptable performance with small mini-batch sizes.

The outline of this paper is as follows: 
Section~\ref{sec:jobs} illustrates the JOBS procedure; shows that it is a relaxation of $\ell_1$ minimization and provide some intuitions for analysis.
Section~\ref{sec:prelim} summarizes 
theoretical background material 
to analyze our algorithm.
Section~\ref{sec:theory} demonstrates all the major theoretical results of JOBS and Bagging with a generic $L/m$ ratio and $K$.
Section~\ref{sec:proofs_main} describes detailed analysis of the results in Section~\ref{sec:theory}.
Finally, Section~\ref{sec:simulation} gives multiple simulation results comparing JOBS, Bagging, Bolasso, as well as $\ell_1$ minimization.

\section{Proposed Method: JOBS} 
\label{sec:jobs}
We first introduce a notation for the general form of the mixed $\ell_{p,q}$ norm of a matrix. The row sparsity penalty that we employed in our proposed method is a special case of this norm with $p = 1, q = 2$. 
The mixed $\ell_{p,q}$ norm on matrix $\X$ is defined as:
\vspace{-0.05in}
\begin{equation}
\label{eq:lpq}
\begin{split}
 \|\X\|_{p,q} & = (\sum_{i = 1}^n \| \xri ^{\transpose} \|^p_q )^{1/p} \\
  & =  \| (\| \xrOne   ^{\transpose} \|_q, \| \xrTwo  ^{\transpose} \|_q,..., \| \xrn^{\transpose} \|_q )^{\transpose} \|_p, 
\end{split}
\end{equation}
where $\x[\ri]$ denotes the $\ri-$th row of matrix $\X$. Intuitively, the mixed $\ell_{p,q}$ norm essentially takes $\ell_q$ norms on rows of $\X$ first; stacks those as a vector and then computes $\ell_p$ norm of this vector. Note when $p = q $, the $\ell_{p,p}$ norm of $\|\X \|$ is simply the $\ell_p$ vector norm on vectorized $\X$. 

\subsection{JOBS}
Our proposed method JOBS can be accomplished in three steps. First, we generate bootstrap samples:
The multiple bootstrap process generates $K$ multi-sets of the original data, which contains $K$ sensing matrices and measurements pairs: $\{ \yIcOne , \AIcOne \} , \{\yIcTwo, \AIcTwo \} ...., \{ \yIcK, \AIcK \} $.
Second, we solve the collaborative recovery on those sets, the optimization in both noiseless and noisy forms.
The noiseless case problem is: 
for all $ \ci   =  1, 2, ..., K $,
\begin{equation}
\label{eq:bmmv_ideal}
\oursNoNoise:   \min   \| \X \|_{1,2} \st    \yIci = \AIci  \xci,
\end{equation}
 and the noisy counterpart can be expressed as: for some $\epsilon^{J}> 0$,
 \vspace{-0.16in}
\begin{equation}
\label{eq:bmmv}
\ours:      \min   \| \X \|_{1,2}   \st \sum_{\ci = 1 }^{K} \| \yIci  - \AIci  \xci  \|_2 \leq \epsilon^{J}.
\end{equation}
 \vspace{-0.1in}

Proposed approaches in $\oursNoNoise$, $\ours$  are in the form of Block(Group) sparse recovery~\cite{spgl1Berg} and numerous optimization methods can solve them such as~\cite{admm,dcs_barron,LOPT,jsprox,glasso,spgl1Berg,spaRSA,liu2009,yall}.

Finally, the JOBS solution is obtained through averaging the columns of the solution of (\ref{eq:bmmv_ideal}) or (\ref{eq:bmmv}): $\Xsharp$, 
\vspace{-0.05in}
\begin{equation}
\mbox{JOBS:} \quad \xf = \frac{1}{K}\sum_{\ci = 1}^K \x^{\boldsymbol{\sharp}}_\ci.
\end{equation}
All supports of $\x^{\boldsymbol{\sharp}}_1, \x^{\boldsymbol{\sharp}}_2, ..., \x^{\boldsymbol{\sharp}}_K$ are the same because of the row sparsity constraint that we impose, and therefore the sparsity of the JOBS solution $\xf$ will not increase as in the Bagging case.

\subsection{Intuitive Explanation of why JOBS Works}
\label{sec:relax}
JOBS recovers the true sparse solution because it is  a relaxation of the original $\ell_1$ minimization problem 
in multiple vectors. 
Let $\optx$ be the true sparse solution; we will show that under some mild conditions, the row sparse minimization program recovers $(\optx, \optx, ..., \optx)$
correctly in Section \ref{sec:theory}. Thus the average over columns returns exactly the true solution $\optx$.

We first demonstrate that JOBS is a
 two-step relaxation procedure of $\ell_1$ minimization. 
 For a $\ell_1$ minimization 
 as in equation (\ref{eq:smv_eq}) with a unique solution $\optx$, the Multiple Measurement Vectors (MMV) equivalence is: 
 for $\ci = 1, 2,..,K$
\begin{equation}
\label{eq:smv_eq_mmv}
 \loneminNoNoiseMulti:  \min \|\X \|_{1,1} \st \y = \A \xci ,
\end{equation}
where $\|\X\|_{1,1} = \sum_i {\|\xri ^\transpose \|_1}.$ 
For the equivalence to the original problem, we have: if Single Vector Measurement (SMV) problem $\loneminNoNoise$  has a unique solution $\optx$,
then the solution to the MMV problem $\loneminNoNoiseMulti$ yields a row sparse solution
$\optX = (\optx, \optx, ..., \optx) $. This result can be derived via contradiction.
The reverse direction is also true: if the MMV problem $\loneminNoNoiseMulti$ has a unique solution, it implies that the SMV problem $\loneminNoNoise$ must also have a unique solution.
One can refer to Lemma~\ref{lemma:p1ktop1} and its proof in Appendix~\ref{app:p1ktop1} for details. 

\begin{figure}[h]
\vspace{-0.15in}
\centering
{\includegraphics[width=0.45\textwidth]{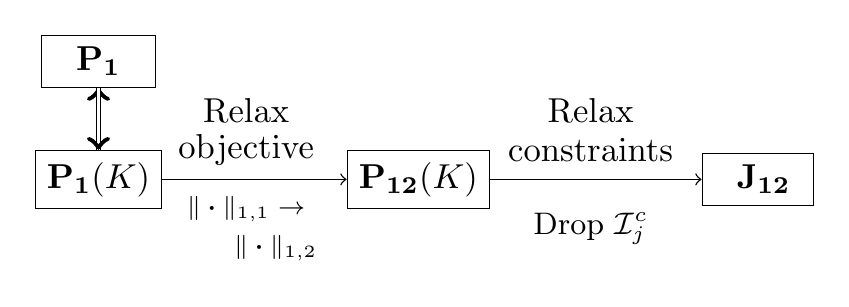}}
\vspace{-0.2in}
\caption{Flowchart explaining JOBS as a relaxation of $\ell_1$ minimization}
\label{fig:flowchart}
\end{figure}

Since the $\ell_{1,1}$ norm is separable for each elements of $\X$, it does not enforce support consistency. We therefore relax the $\ell_{1,1}$ norm in (\ref{eq:smv_eq_mmv}) to the $\ell_{1,2}$ norm. 
For all $ \ci  =  1,2,.., K$ 
\begin{equation}
\label{eq:mmv_re}
\lTwoOneNoNoise: \ \min \|\X \|_{1,2} \st \y = \A \xci. 
\end{equation}
To obtain $\oursNoNoise$ in (\ref{eq:bmmv_ideal}), for each $\xci$, we further relax the problem by dropping all constraints that are not in $\I_\ci$ from (\ref{eq:mmv_re}). This two-step relaxation process is illustrated
in Figure~\ref{fig:flowchart}.

The noisy version can be analyzed similarly. We can formulate 
the  MMV version of the original problem; relax the regularizer from $\ell_{1,1}$ norm to $\ell_{1,2}$  norm, and then further relax the objective function by dropping the constraints that are not on the selected subset $\I_j$ for $j-$th estimate $\xci$ to obtain the
proposed form $\ours$.
\vspace{-0.05in}
\vspace{-0.05in}

\section{Preliminaries}
\label{sec:prelim}

We summarize the theoretical results that are needed for understanding and analyzing our algorithm mathematically.  We are going to introduce block sparsity, Null Space Property (NSP), as well as Restricted Isometry Property (RIP).
\subsection{Block (non-overlapping group) Sparsity}
Because row sparsity
is a special case of block sparsity (or non-overlapping group sparsity),
we therefore can employ the tools from block sparsity to analyze our problem. To start, recall the definition of block sparsity as in~\cite{blocksparse}:

\begin{definition}[Block Sparsity]
$\x \in \R^n$ is $\sparse-$block sparse with respect to a partition $\B = \{  \B_1, \B_2, ..., \B_K \}$ of $\{1,2,..., n \}$ if
$\x = (  \x{\text{\scriptsize{$[\B_1]$}}} , ....  \x{\text{\scriptsize{$[\B_K]$}}} )$, the norm $\| \x \|_{2, 0 | \B} : = \sum_{i =1}^K \indicator \{ \| \x{\text{\scriptsize{$[\B_i]$}}}  \|_2  > 0 \} \leq \sparse $
and the relaxation $\ell_{1,2}$ norm $\| \x \|_{2, 1 | \B} : = \sum_{i = 1}^{K}   \| \x{\text{\scriptsize{$[\B_i]$}}} \|_2 $.
\end{definition}

The block sparsity level $\| \x \|_{2, 0 | \B} $ counts the number of non-zero blocks of the given block partition $\B$. Block sparsity is also a generalization of standard sparsity.
Usually, for the same sparse vector $\x$, the group sparsity level is smaller than the sparsity level. Therefore knowing the group sparse information may reduce the number of minimum measurements needed comparing to standard sparse recovery.

We can see that $\ell_{1,2}$ minimization is a special case of block sparse minimization~\cite{blocksparse}, with each element in the group partition containing the indices of each row. Therefore, the analysis of our algorithm follows similar analyses in the studies of block sparsity such as Block Null Space Property (BNSP) \cite{bnsp}, Block Restricted Isometry Property (BRIP) \cite{brip}.

\subsection{Null Space Property (NSP) and Block-NSP (BNSP) }
The NSP~\cite{csNSP} for standard sparse recovery and block sparse signal recovery are given in the two following theorems.
\begin{theorem}[NSP]
\label{nsp_smv}
Every $\sparse-$sparse signal $\x \in \R^{n }$ is a unique solution to
$\loneminNoNoise : \ \min \| \x \|_{1} \st \y = \A \x $
if and only if $\A$ satisfies NSP of order $\sparse$.
Namely, if for all $\v \in \Null{(\A)} \backslash \{ \zero \}$, such that for any set $\S$ of cardinality less than equals to $\sparse$
$ : \S \subset \{1,2,.., n \}, \text{card}(\S) \leq \sparse $, the following is satisfied:
\begin{equation*}
\|  \v \text{\footnotesize{$[\S]$ }} \|_{1} < \| \v \text{\footnotesize{$[\S^c]$ }} \|_{1},
\end{equation*}
where  $\v \text{\footnotesize{$[\S]$ }}$ only has the vector values
on a index set $\S$ and zero elsewhere.
\end{theorem}

\begin{theorem}[BNSP]
\label{thm:bnsp}
Every $\sparse-$block sparse signal $\x$ respect to block assignment $\B$, $\A$ is unique solution to $\ \min \| \x \|_{1,2|\B} \st \y = \A \x $ if and only if $\A$ satisfies Block Null Space Property (BNSP) of order $\sparse$:\\
For any set $\S \subset \{ 1, 2, ..., n \} $ with card$(\S) \leq \sparse$, a matrix $\A$ is said to satisfy 
block null space property over $\B $ of order $\sparse$, if
\begin{equation*}
\|\v \text{\footnotesize{$[\S]$ }} \|_{1,2|\B} < \| \v \text{\footnotesize{$[\S^c]$ }} \|_{1,2|\B},
\end{equation*}
for all $\v \in \Null{(\A)} \backslash \{ \zero \}$, where  $\v_\S$ denotes the vector equal to  $\v$ on a block index set $\S$ and zero elsewhere.
\end{theorem}

\subsection{Restricted Isometry Property (RIP) and Block-RIP (BRIP) }
Although NSP directly characterizes the ability of success for sparse recovery, checking the NSP condition is computationally intractable, and it is also not suitable to use NSP for quantifying performance in noisy conditions since it is a binary (True or False) metric instead of in a continuous range. Restricted isometry property (RIP) is introduced for those purposes and there are many sufficient conditions based on RIP. Let us recall RIP~\cite{cs} for standard sparse recovery and BRIP~\cite{brip} for block sparse recovery.
\begin{definition}[RIP]
\label{def:rip}
A matrix $\A$ with $\ell_2$-normalized columns satisfies RIP of order $\sparse$ if there exists a constant $\delta_{\sparse}(\A) \in [0 , 1) $ such that for every $\sparse-$sparse $\v \in \R^n$, we have:
\begin{equation}
\label{eq:def_rip}
(1 - \delta_{\sparse} (\A) ) \|  \v \|_2^2 \leq  \| \A \v \|_2^2  \leq ( 1 + \delta_{\sparse}(\A) ) \| \v \|_2^2.
\end{equation}
\end{definition}

More generally, the RIP condition for block sparsity definitions (Definition 2 in \cite{brip}) are as the following:
\begin{definition}[BRIP]
\label{def:def_brip}
A matrix $\A$ with $\ell_2$-normalized columns satisfies Block RIP with respect to block partition $\B$ of order $\sparse$ if there exists a constant $\delta_{ \sparse | \B} (\A) \in [0 , 1) $ such that for every $\sparse-$block sparse $\v \in \R^n$ over $\B$, we have:
\begin{equation}
\label{eq:brip}
(1 - \delta_{ \sparse | \B} (\A) ) \|  \v \|_2^2 \leq  \| \A \v \|_2^2  \leq ( 1 + \delta_{ \sparse | \B}(\A) ) \| \v \|_2^2.
\end{equation}
\end{definition}

Again, if we take every entry as a block, the block sparsity RIP reduces to the standard RIP condition. 

%

\subsection{Noisy Recovery bounds based on RIP constants}
It is known that RIP conditions imply NSP conditions satisfied for both block sparse recovery and sparse recovery. More specifically, if the RIP constant in the order $2\sparse$ is strictly less than $\sqrt{2} - 1 $, then it implies that NSP is satisfied in the order of $\sparse$. This applies to both classic $\ell_1$ sparse recovery and block sparse recovery.

The noisy recovery performance bound based on RIP constant for $\ell_1$ minimization problem 
and the noisy recovery bound for block sparse recovery based on BRIP constant are shown in the following two theorems.
\begin{theorem}[Noisy recovery for $\ell_1$ minimization, Theorem 1.2 in~\cite{cs}]
\label{th:noisy_recon_l1}
Let $ \y = \A \optx + \z$, $ \| \z \|_2 \leq \epsilon$, $\x_0$ is $\sparse-$sparse that minimizes $\| \x - \optx \|$ over all $\sparse-$ sparse signals. If $\delta_{2 \sparse } (\A) < \sqrt{2} - 1$, $\x^{\boldsymbol{\ell_1}} $ be the solution of $\ell_{1} $ minimization
, then
\begin{equation*}
\| \x^{\boldsymbol{\ell_1}} - \optx \|_2 \leq \cOne(\delta_{2 \sparse } (\A)) \sparse^{-1/2} \| \x_0 - \optx\|_{1} + {\cTwo(\delta_{2 \sparse} (\A))} \epsilon ,
\end{equation*}
where $ \cOne(\cdot), \cTwo(\cdot)$ are some constants, which are determined by RIP constant $\delta_{2 \sparse}$. The form of these two constants terms are $\cOne(\delta) = \frac{ 2( 1 - (1 - \sqrt{2} )\delta)} {1- (1+ \sqrt{2})\delta }$ and $\cTwo (\delta) =  \frac{ 4 \sqrt{1 + \delta }}{ 1 - ( 1 + \sqrt{2} )  \delta }$.
\end{theorem}

\begin{theorem}[Noisy recovery for block sparse recovery, Theorem 2 in~\cite{blocksparse}]
\label{th:noisy_recon}
Let $ \y = \A \optx + \z$, $ \| \z \|_2 \leq \epsilon$, $\x_0$ is $\sparse-$block sparse that minimizes $\| \x - \optx \|$ over all $\sparse-$block sparse signals. If $\delta_{2 \sparse |\B} (\A) < \sqrt{2} - 1$, 
$\x^{\boldsymbol{\ell_{1,2|\B}}}$ be the solution of block sparse minimization, then
\begin{equation*}
\begin{split}
\| \x^{\boldsymbol{\ell_{1,2|\B}}} - \optx \|_2 & \leq \cOne(\delta_{2 \sparse |\B}(\A)) \sparse^{-1/2} \| \x_0 - \optx\|_{1,2|\B} \\
 & + {\cTwo(\delta_{2 \sparse |\B}(\A))} \epsilon ,
\end{split}
\end{equation*}
where $ \cOne(\cdot), \cTwo(\cdot)$
are the same functions as in Theorem~\ref{th:noisy_recon_l1}.
\end{theorem}

\subsection{Sufficient Condition: Sample Complexity for Gaussian and Bernoulli Random Matrices}
Since checking either NSP or RIP conditions is computationally hard and it doesn't provide direct guidance for designing sensing matrices, some previous work built a relationship between sample complexity for random matrices to a designed RIP constant. The classical one is Theorem 5.2 in \cite{simpleRIP}:
\begin{theorem}[Sufficient Condition: Sample Complexity]
\label{th:rdmtx}
Let entries of $ \A \in \R^ {m \times n}$ from $\mathcal{N}(0, 1/m ), 1/\sqrt{m}$ Bern($0.5$). Let $\mu , \delta \in ( 0,1) $ and assume $m \geq \Const \delta^{-2}(\sparse \ln(n/ \sparse ) + \ln(\mu^{-1}))$ for a universal constant $\Const > 0$, then $\P (\delta_\sparse (\A) \leq \delta) \geq 1 - \mu $.
\end{theorem}

By rearranging the terms in this theorem, the sample complexity can be derived: when $m$ is in the order of $\O( 2 \sparse  \ln ( n / 2\sparse ) )$ and sufficient large, there is a high probability that the RIP constant of order $2 \sparse $ is sufficiently small.


\section{Theoretical Results for JOBS}
\label{sec:theory}

\subsection{BNSP}
Similarly to previous CS analysis in~\cite{cs}, we give the null space property to characterize the exact recovery condition of our algorithm.  The BNSP for JOBS is stated as follows:
\begin{definition}[BNSP for JOBS]
\label{th:nsp_gmmv}
A set of sensing matrices $\{ \A_1, \A_2,..., \A_K \}$ satisfies BNSP of order $\sparse$ if
$ \ \forall \ ( \v_1 , \v_{2}, ..., \v_{K}) \in \Null (\A_1)  \times \Null(\A_2) ... \times \Null(\A_K) \backslash \{ (\zero, \zero, ..., \zero) \}$, such that for all $\S : \S \subset \{1, 2, ..., n\}, \text{card}(\S) \leq \sparse $: 
 $\|  \V [ \S ] \|_{1,2} < \| \V [\S^c] \|_{1,2}$.
\end{definition}
\begin{theorem}[Necessary and Sufficient Condition for JOBS]
\label{th:jobs_noiseless}
(i) $\oursNoNoise$ successfully recovers all the $\sparse-$row sparse solution if and only if $\{\A{[\I_1]}, \A{[\I_2]},..., \A{[\I_K]}
\}$ satisfies BNSP of the order of $\sparse$. 
(ii) The solution is of the form $\optX = (\optx, \optx,...,\optx)$, where $\optx$ is the unique solution to $\loneminNoNoise$. Then, the JOBS solution $\xf$ is the average over columns of $\optX$, which is
$\optx$.
\end{theorem}

 Obtaining Definition~\ref{th:nsp_gmmv} is straight forward. We prove it using the
 BNSP of the general $\ell_{p,2}$ block norm stated in Appendix in Definition~\ref{p2_block_min}. 
Theorem~\ref{th:jobs_noiseless} {\it (i)} can be obtained from BNSP in~\cite{blocksparse}
and Theorem~\ref{th:jobs_noiseless} {\it (ii)} can be derived by showing that $\optX$ is feasible and it achieves the lower bound of $\ell_{1,2}$ norm of feasible solutions.
The BNSP of JOBS characterizes the existence and unique of solution and Theorem~\ref{th:jobs_noiseless} establishes the correctness of JOBS.

\subsection{BRIP}

Since the BNSP is in general difficult to check, RIP, a more applicable quantity is derived.
 It useful in analyzing the error bounds for the noisy cases, where both sufficient conditions and error bounds are related to the RIP constant.
We will show that the BRIP constant for JOBS can be decomposed to the maximum of RIP constants for all sensing matrices.

Let $\AJ = \text{block}\_\diag( \A{[\I_1]} ,\A{[2]}, ..., \A{[\I_K]})$ 
and $\B = \{\B_1, \B_2, ..., \B_n \}$ be the group partition of $\{1, 2,..., nK\}$ that corresponds to row sparsity pattern.
 Let $\delta_{ \sparse|\B}$ denote row sparse BRIP constant of order $\sparse$ 
 and $\delta_{\sparse}$ denote RIP constant of order $\sparse$. The BRIP constant for JOBS is as follows.

\begin{proposition}[BRIP for JOBS]
\label{prop:factorization}
For all $\sparse \leq  n , \sparse \in \mathbb{Z}^+ $ 
\begin{equation}
\label{eq:oursRIP}
\delta_{\sparse| \B} (\AJ) = \max_{ \ci = 1,2,..., K } \delta_{\sparse} (\A{[\I_\ci]} ) .
\end{equation}
\end{proposition}
The proof of this proposition is elaborated in Appendix~\ref{pf:prop_factorization}. It is not so surprising that the BRIP of JOBS depends on the worst case among all $K$ choices of sub-matrices since smaller RIP constant indicates better recovering ability.
 More importantly, this result shows that the BRIP constant for JOBS can be decomposed into functions of standard RIP constant for each sub-matrix, which enables us to derive the sample complexity of our algorithm simply based on the sample complexity for $\ell_1$ minimization
in Theorem~\ref{th:rdmtx}.  

\subsection{Noisy Recovery Performance}

From previous analyses we can establish that if the BRIP constant of order
$2\sparse$ is less than $\sqrt{2}-1$, it implies that $\{\A{[\I_1]}, \A{[\I_2]},..., \A{[\I_K]}\}$ satisfies BNSP of order $\sparse$. Then, by Theorem~\ref{th:jobs_noiseless}~{\it (ii)}, we know that the optimal solution to $\oursNoNoise$ is the $\sparse-$row sparse signal $\optX$ with every column being $\optx$. Similar to block sparse recovery bound in Theorem \ref{th:noisy_recon}, the reconstruction error is determined by the $\sparse-$block sparse approximation error and the noise level.  In the case when the true solution is exactly $\sparse-$row sparse, it is relatively easy to analyze. For each realization, its performance can be analyzed through Theorem 2 in~\cite{blocksparse}. Then to characterize typical performance of JOBS, we use the tail bounds 
and obtain the following result.

\begin{theorem}[JOBS: Error bound for $\|\optx \|_0 = \sparse$ ]
\label{th:noisy_recon_ours_exact_sparse}
Let $ \y = \A \optx + \z$, $ \| \z \|_2 < \infty $. If $\delta_{2 \sparse| \B}(\AJ) \leq \delta < \sqrt{2} -1$ and the true solution is exactly $ \sparse-$sparse, then for any $\tau>0$, JOBS solution $\xf$ satisfies
\begin{equation}
\label{eq:ours}
\begin{split}
\P & \{ \| \xf - \optx \|_2      \leq  \cTwo(\delta_L) ( \sqrt{ \frac{L}{m} } \| \z \|_2 + \tau   )    \}  \\
&\geq 1 -  \exp{\frac{-2 K \tau^4  }{ L \|\z\|_\infty^4}}.
\end{split}
\end{equation}
\end{theorem}


In the more general case, when the sparsity level of $\optx$ possibly exceeds $\sparse$,
there is no guarantee that the non $\sparse-$sparse part will be preserved by JOBS relaxation. Namely, let $\Xwhatever$ denote the true solution for the noiseless row sparse recovery program: $\oursNoNoise$, if BNSP of order greater than $\sparse$ is not guaranteed to be satisfied, then it is not guarantee that $\Xwhatever = \optX$.  
However, if $\optx$ is a near $s-$sparse, then $\Xwhatever $ is not far away from $\optX$. Since $\Xsharp$, recovered from $\ours$, is close to $\Xwhatever$ via the block sparse recovery bound, 
 $\Xsharp$ is close enough to $\optX$. This result is stated in the following theorem.

\begin{theorem}[JOBS: Error bound for general signal recovery]
\label{th:noisy_recon_ours}
Let $ \y = \A \optx + \z$, $ \| \z \|_2 < \infty $,
If $\delta_{2 \sparse|\B}(\AJ) \leq \delta < \sqrt{2} -1$,
then for any $\tau > 0$, JOBS solution $\xf $ satisfies
\begin{equation}
\label{eq:errBound}
\begin{split}
& \P \{ \| \xf - \optx \|_2   \leq \|\e\|_2+ \cTwo(\delta_L) ( \sqrt{\frac{L}{m}} \| \A\e+ \z \|_2
 + \tau )  \} \\
&\geq  1 - \exp \frac{-2 K \tau^4}{L (  \| \A\|_{\infty,1} \|\e\|_\infty  + \| \z\|_\infty)^4},
\end{split}
\end{equation}
where $\e$ is the $\sparse$-sparse approximation error: $\e = \optx -\x_0 $ with $\x_0$ being the top $\sparse$ components of the true solution $\optx$, and $\|\A\|_{\infty,1}=
\max_{ \ri = 1, 2, ..., m} ( \| \ari^\transpose  \|_1) $ denotes the largest $\ell_1$-norm of all rows of $\A$.
\end{theorem}


The error bound in Theorem~\ref{th:noisy_recon_ours} relates to $\sparse-$sparse approximation error as well as the noise level, which is similar to $\ell_1$ minimization and block sparse recovery bounds. JOBS also introduces a relaxation error bounded by $\| \e \|_2$. The smaller the power of $\e$, the smaller the upper bound. When $\e = \zero$, $\optx$ is exactly $\sparse-$sparse, then Theorem~\ref{th:noisy_recon_ours} reduces to Theorem~\ref{th:noisy_recon_ours_exact_sparse}~.
From those two theorems, there are trade-offs for a good choice of 
bootstrap sample size $L$ and number of bootstrap samples $K$. The relationship of the bound and $L$ is the following:
Because the BRIP constant decreases with the increasing $L$ and $\cTwo(\delta)$ is a non-decreasing function of $\delta$, a larger $L$ results in a smaller $\cTwo(\delta)$. The ratio $\sqrt{L/m}$, however, is smaller for smaller $L$.
As for the number of estimates, the uncertainty in (\ref{eq:errBound}) decays exponentially with $K$, so a large $K$ is preferable in this sense.

\subsection{Noisy Recovery for Employing Bagging in Sparse Recovery}
We also derive the performance bound for employing Bagging scheme to sparse recovery problem, 
in which the final estimate is the average over multiple estimates solved individually from bootstrap samples. We give the theoretical results for the case that true signal $\optx$ is exactly $\sparse-$sparse and the general case that it is not necessarily exactly $\sparse-$sparse.
\begin{theorem}[Bagging: Error bound for $\|\optx \|_0 = \sparse $ ]
\label{th:noisy_bagging_exact_sparse}
Let $ \y = \A \optx + \z$, $ \| \z \|_2 < \infty $, 
If under the assumption that, for $\{ \I_\ci\}$s that generates a set of sensing matrices $\A{[\I_1]}, \A{[\I_2]},..., \A{[\I_K]}$,
there exists $\delta$ such that for all $\ci \in \{1,2,...,K\}$, $\delta_{2 \sparse }(\A{[\I_\ci]}) \leq \delta < \sqrt{2} -1$. Let
$\xB $ be the solution of Bagging, then
for any $\tau > 0$, $\xB$ satisfies
\begin{equation}
\label{eq:sum_upbd1}
\begin{split}
  \P    & \{   \| \xB- \optx \|_2  \leq
 \cTwo(\delta_L)  (  \sqrt{ \frac{L}{m}}  \| \z \|_2 + \tau)  \}\\
   & \geq 1 - \exp \frac{- 2 K  \tau^4   }{ L^2 \| \z\|^4_\infty  } .
\end{split}
\end{equation}
\end{theorem}


We also study the behavior of Bagging for general signal $\optx, \| \optx \|_0 \geq \sparse $, in which the performance involves the $\sparse-$sparse approximation error.
\begin{theorem}[Bagging: Error bound for general signal recovery]
\label{th:noisy_bagging_dm}
Let $ \y = \A \optx + \z$, $ \| \z \|_2 < \infty $,
If under the assumption that, for $\{ \I_\ci\}$s that generates a set of sensing matrices $\A{[\I_1]}, \A{[\I_2]},..., \A{[\I_K]}$, there exists $\delta$ such that for all $\ci \in \{1,2,...,K\}$, $\delta_{2 \sparse }(\A{[\I_\ci]}) \leq \delta < \sqrt{2} -1$. Let $\xB $ be the solution of Bagging, then
for any $\tau > 0$, $\xB$ satisfies
\begin{equation}
\begin{split}
  \P & \{  \| \xB- \optx\|_2 \leq
(\cOne(\delta_L) \sparse^{-1/2} \| \e \|_{1} + {\cTwo(\delta_L)} (\sqrt{ \frac{L}{m}} \| \z \|_2 + \tau )  \} \\
& \geq 1 - \exp  \frac{ - 2 K \cTwo^4(\delta) \tau^4}{ (b'    )^2} ,
\end{split}
\end{equation}
where $b' = (\cOne(\delta) \sparse^{-1/2} \| \e\|_{1} + {\cTwo(\delta)} \sqrt{L}\| \z \|_\infty)^{2} $.
\end{theorem}
Theorem~\ref{th:noisy_bagging_dm} gives the performance bound for Bagging for general signal recovery without the $\sparse-$sparse assumption, and it reduces to Theorem~\ref{th:noisy_bagging_exact_sparse} when the $\sparse-$sparse approximation error is zero $\|\e\|_1=0$.   
Both Theorem~\ref{th:noisy_bagging_exact_sparse} and~\ref{th:noisy_bagging_dm} above show that increasing the number of estimates $K$ improves the result, by increasing lower bound of the certainty of the same performance.

Here are some comments about the error bound for JOBS compared to Bagging. The RIP condition for Bagging is the same as the RIP condition for our algorithm, under the assumption that all submatrices $\A{[\I_{\ci}]}$ are well-behaved. When $\|\optx\|_0 = \sparse$, the bound in Bagging is worse than JOBS, since the certainty for algorithm is at least $ 1 - \exp  \frac{- 2 K  \tau^4 }{ L^2\| \z\|^4_\infty}$, compared to the error bound  $ 1 - \exp  \frac{-2 K  \tau^4 }{ L\| \z\|^4_\infty}$ in JOBS. With a squared $L$ instead of $L$, that term is smaller than the term in JOBS for the same choices of $L$ and $K$.

As for the general signal recovery bound of Bagging in Theorem~\ref{th:noisy_bagging_dm}, since the error bound for bagging does not contain the MMV relaxation error as the one in JOBS, however the tail bound involves more complicated terms. This bound is nontrivial comparing to the one to JOBS. Although the $\sparse-$sparse assumption limits to exact $\sparse-$sparse signal, for signals that are approximately $\sparse-$sparse, or with low energy in the $\sparse-$sparse approximation ($\|\e\|_1$ is low), the behavior would be close to the exact $\sparse-$sparse case.


\subsection{Sample Complexity for JOBS with i.i.d. Gaussian Sensing Matrices}
\begin{theorem}[Sample Complexity for JOBS]
\label{cor:suff} If the entries of the original sensing matrix $\A$ are i.i.d Gaussian or sub-Gaussian, then
for $d < L$, a small $\alpha$ and a small $\mu>0 $,  such that the minimum number of distinct elements across $\I_i$s are bounded by $d$:
$\P \{  V(\I_i) \geq d  \} \geq 1 - \alpha$ ($V(\I)$ counts the number of distinct elements in multi-set $\I$),
and if $ d $ is in the order of $\O(2 \sparse \ln(n/ 2 \sparse) + \ln K   +\ln( \frac{(1 - \alpha)^K}{(1 - \alpha)^K - (1 - \mu)}))$, and the constant depending on $\alpha, \mu$ and $K$, $\P ( \delta_{2 \sparse|\B} ( \AJ  )  <  \sqrt{2} - 1 ) \geq 1 - \mu$, and therefore, JOBS recovers the true $\sparse-$sparse solution $\optX$ with at least a certain probability relates to $K, \alpha, \mu$.
\end{theorem}

In the sample complexity analysis, the $2 \sparse \ln(n/2\sparse)$ term coincides with the one from $\ell_1$ minimization. 
The terms associated with $K $ are non-decreasing with respect to $K$, which is introduced by the increasing uncertainty from taking a large number of bootstrap samples, resulting from the non-decreasing property of BRIP with adding extra sets of bootstrap samples shown in
(\ref{eq:oursRIP}).


\section{Proofs of main theoretical results}
\label{sec:proofs_main}
%
%
%


\subsection{Proof of Necessary and Sufficient condition: Theorem~\ref{th:jobs_noiseless}}
Theorem~\ref{th:jobs_noiseless} {\it (i)} can be directly shown from the BNSP for block sparse minimization problems as in~\cite{blocksparse}.

We show the procedure to prove Theorem~\ref{th:jobs_noiseless} {\it (ii)}. If BNSP of order $\sparse$ is satisfied for $\{ \A{[\I_1]}, \A{[\I_2]}, ..., \A{[\I_K]} \}$, then each submatrix $\A{[\I_\ci]}$ satisfies Null Space Property (NSP) of order $\sparse$. The detailed proof is in Appendix \ref{app:p1ktop1} in proving Lemma \ref{lemma:p1ktop1}.
%
%
%
%
Consequently, for all $\ci = 1,2,...,K $, let $\optx$ be the optimal solution:
\begin{equation}
\optx = \argmin_{\xci}  \| \xci \|_1 \st    \yIci=  \AIci \xci .
\end{equation}

For $\X$ a feasible solution, consider its $\ell_{1,2} $ norm, we have:
\begin{equation*}
\begin{split}
 \|\X\|_{1,2}  & = \sum_{ \ri = 1}^n ( \sum_{\ci = 1}^K ( x_{\ri \ci}^2 ))^{1/2} = \sqrt{K} \sum_{\ri = 1}^n  (\frac{1}{K}\sum_{\ci = 1}^K ( x^2_{ \ri \ci } ))^{1/2}
\end{split}
 \end{equation*}
By concavity of square root, we have
\begin{equation*}
\begin{split}
& \geq \sqrt{K} \sum_{\ri = 1}^n \frac{1}{K} \sum_{ \ci = 1}^K \sqrt{ x^2_{ \ri \ci} } = \sqrt{K}  \frac{1}{K} \sum_{\ci = 1}^K \sum_{\ri = 1}^n |x_{ \ri \ci }| \\
& \geq \sqrt{K} \frac{1}{K} \sum_{\ci = 1}^K \min_{ \substack{ \xci: x_{1\ci},...,x_{n\ci}  \\  \AIci \xci = \yIci } } \sum_{\ri = 1}^n |x_{ \ri \ci}| \\
& = \sqrt{K} \frac{1}{K} \sum_{\ci = 1}^K \min_{\xci : \AIci \xci = \yIci } \| \xci \|_1 \\
& = \sqrt{K} \|\optx\|_1.
\end{split}
 \end{equation*}
$ \optX =  (\optx, \optx,..., \optx) $ is a feasible solution and $\| \optX \|_{1,2} = \|( \optx, \optx,..., \optx) \|_{1,2} = \sqrt{K} \|\optx\|_1$, and it achieves the lower bound. By uniqueness from (i), we can concluded that $\optX$ is the unique solution. 
The JOBS solution takes average over columns of multiple estimates. Since the average of $\optX$ is $\optx$, we prove that JOBS returns the correct answer.


\subsection{Proof of Theorem \ref{th:noisy_recon_ours_exact_sparse}: performance bound of  for exactly $\sparse-$sparse }

If the true solution is exactly $\sparse-$sparse, the sparse approximation error is zero. Then the noise level of performance only relates to measurements noise.
For $\ell_1$ minimization, $\z$ is the noise vector and we use matrix $\Z = (\zIcOne, \zIcTwo,..., \zIcK)$ to denote the noise matrix in JOBS. We bound the distance of $\| \Z \|_{2,2}$ to its expected value using Hoeffidings' inequalities stated in~\cite{hoeffding1963} by Hoeffding in 1963.
%

\begin{theorem}[Hoeffdings' Inequalities]
Let $X_1, ...,X_n$ be independent bounded random variables such that $X_i$ falls in the interval $[a_i, b_i]$ with probability one. Denote their sum by $S_n = \sum_{i = 1}^n X_i$. Then for any $\epsilon > 0 $, we have:
\begin{equation}
\label{eq:hoeff1}
\P \{ S_n - \E \S_n \geq \epsilon \} \leq \exp{\frac{-2 \epsilon^2}{ \sum_{i=1}^n (b_i - a_i )^2 }} \ \ \mbox{and}
\end{equation}
\begin{equation}
\P \{ S_n - \E \S_n \leq -\epsilon \} \leq \exp{\frac{-2 \epsilon^2}{ \sum_{i=1}^n (b_i - a_i )^2 }}.
\end{equation}
\end{theorem}

Here, the whole noise vector is $\z = \A \x - \y = ( \zrOne, \zrTwo, ..., \zrm)^T, \|\z\|_\infty = \max_{i \in \{ 1, 2, ..., m\}} | z \text{\footnotesize{$[i]$}} | < \infty$. We consider the matrix  $\Z \circ \Z = (\xi_{ji})$, where $\circ$ is the entry-wise product. The quantity that we are interested in $\|\Z\|_{2,2}$ is the sum of all entries in $\Z \circ \Z$. Each element in this matrix $\Z \circ \Z$
 is drawn i.i.d from the squares of entries in $\z$: 
 $\{\zrOne, \zrTwo, ..., \zrm\}$ with equal probability. 
Let $\Xi$ be the underlining random variable 
and $\Xi$ obeys a discrete uniform distribution
\begin{equation}
\P( \Xi = z^2 \text{\footnotesize{$[i]$}}) = \frac{1}{m} , i = 1, 2,..., m. 
\end{equation}
The lower and upper bound of $\Xi$ is
\begin{equation}
0 \leq \min_{i } z^2[i] \leq \Xi  \leq   \|\z\|^2_\infty. 
\end{equation}
 We use zero as lower bound for $\Xi$ instead of the minimun value to simplify the terms. 
The expected power of $\Z$  is 
  \begin{equation}
  \label{eq:exp_noise}
   \E \| \Z  \|_{2,2}^2  = \frac{KL}{m} \| \z \|_2^2.
   \end{equation}


Then applying Hoeffdings' inequality (\ref{eq:hoeff1}), 
for any $\tau > 0 $, we have:
\begin{equation}
\label{eq:z_ez}
\P \{ \|\Z  \|^2_{2,2} - \E \|\Z\|^2_{2,2}  - \tau \leq 0 \} \geq 1 - \exp{\frac{-2 \tau^2}{KL \|\z\|_\infty^4}}.
\end{equation}

Let $\Xsharp$ be the solution of $\ours$, and by Theorem \ref{th:noisy_recon} : 
\begin{equation}
\label{eq:bd_ours}
\P \{ \| \Xsharp - \optX\|^2_{2,2} - \mathcal{C}_1^2(\delta) \| \Z\|^2_{2,2} \leq 0 \} = 1. 
\end{equation}
Let $\diffX$ be the difference between the solution to the truth solution scaled by $\cTwo$ constant:
$\diffX = \frac{1}{\cTwo(\delta)}\| \Xsharp - \optX\|_{2,2}$ and (\ref{eq:bd_ours}) becomes
\begin{equation}
\label{eq:bd_ours_simp}
\P \{ \diffX -  \| \Z\|_{2,2} \leq 0 \} = 1. 
\end{equation}
Since $\Z$ depends on the choice of $\I_1, \I_2, ..., \I_K$, we derive the typical performance by studying the distance of the solution to the expected noise level of JOBS.
\begin{equation*}
\label{eq:jobs_noiselevel}
\begin{split}
 & \P   \{ \diffX^2 -  \E \|\Z\|^2_{2,2} - \tau^2 \leq 0 \}   \\
  & = \P\{ \diffX^2 - \|\Z\|^2_{2,2} + \| \Z \|^2_{2,2} - \E \|\Z\|^2_{2,2}  - \tau^2 \leq 0 \}  \\
 & \geq \P\{   \diffX^2 - \|\Z\|^2_{2,2} \leq 0 ,    \| \Z \|^2_{2,2} - \E \|\Z\|^2_{2,2}  - \tau^2 \leq 0  \}\\
 & \quad \mbox{(The first and the second parts are independent)}\\
  & =  \P\{   \diffX^2 - \|\Z\|^2_{2,2} \leq 0 \} \P\{  \| \Z \|^2_{2,2} - \E \|\Z\|^2_{2,2}  - \tau^2 \leq 0  \}  \\
  & \quad \mbox{(using (\ref{eq:bd_ours_simp}) and (\ref{eq:z_ez}))}\\
  &   \geq 1 - \exp{\frac{-2 \tau^4}{KL \|\z\|_\infty^4}}.
\end{split}
\end{equation*}
This procedure gives:
\begin{equation}
\label{eq:show_bd}
  \P   \{ \diffX^2 \leq  \E \|\Z\|^2_{2,2} + \tau^2  \}   \geq 1 - \exp{\frac{-2 \tau^4}{KL \|\z\|_\infty^4}}.
\end{equation}
We bound the squared error as the following: 
\begin{equation}
\label{eq:noiselevel}
\begin{split}
\P& \{ \diffX \leq (\E \|\Z\|^2_{2,2})^{1/2} + \tau \}\\
= \P & \{ \diffX^2 \leq \E \|\Z\|^2_{2,2} + \tau^2 + 2 \tau (\E \|\Z\|^2_{2,2})^{1/2} \} \\
 \geq \P &  \{ \diffX^2  \leq \E \|\Z\|^2_{2,2} + \tau^2  \} . 
\end{split}
\end{equation}
Combining (\ref{eq:show_bd}) and (\ref{eq:noiselevel}), we have:
\begin{equation}
\label{eq:noiselevel_con}
\P \{ \diffX \leq (\E \|\Z\|^2_{2,2})^{1/2} + \tau \} \geq 1 - \exp{\frac{-2 \tau^4}{KL \|\z\|_\infty^4}}.
\end{equation}
Since $f(\x) = \| \x - \optx \|_2^2 $ is convex, we can apply Jensens' inequality: 
\begin{equation}
\label{eq:jensen_l22}
\| \frac{1}{K} \sum_{\ci = 1}^K \x^{\boldsymbol{\sharp}}_\ci - \optx \|^2_2 \leq   \frac{1}{K} \sum_{\ci =1}^K
 \|\x^{\boldsymbol{\sharp}}_\ci - \optx  \|^2_2 .
\end{equation}
The JOBS estimate is averaged over columns of all estimates: 
$\xf  = \frac{1}{K} \sum_{\ci = 1}^K \x^{\boldsymbol{\sharp}}_\ci$.
Therefore, equation (\ref{eq:jensen_l22}) is essentially 
\begin{equation}
\label{eq:jensen}
\P \{ \| \xf - \optx \|^2_2 - \frac{1}{K} \| \Xsharp - \optX\|^2_{2,2} \leq 0 \} = 1.
\end{equation}

Now we consider the typical performance of the JOBS solution:
\begin{equation}
\begin{split}
\P  &\{ \| \xf - \optx \|_2   -  \frac{\cTwo(\delta) }{\sqrt{K} }  ((\E \|\Z\|_{2,2}^2)^{1/2} + \tau  )    \leq 0 \}\\
= & \P \{ \| \xf - \optx \|_2    -   \frac{1}{\sqrt{K}}\| \Xsharp - \optX\|_2 \\
& + \frac{1}{\sqrt{K}}\|\Xsharp - \optX\|_2 - \frac{\cTwo(\delta) }{\sqrt{K} } ( (\E \|\Z\|_{2,2}^2)^{1/2} + \tau  ) \leq 0 \}   \\
\geq  & \P  \{ \| \xf - \optx \|_2   -    \frac{1}{\sqrt{K}}\| \Xsharp - \optX\|_2   \leq 0 ,
\\ & \diffX \leq (\E \|\Z\|^2_{2,2})^{1/2} + \tau  \}\\
  =  &  \P  \{ \| \xf - \optx \|_2   -    \frac{1}{\sqrt{K}}\| \Xsharp - \optX\|_2   \leq 0  \}\\
   & \P \{ \diffX \leq (\E \|\Z\|^2_{2,2})^{1/2} + \tau   \}
\qquad \mbox{(by (\ref{eq:jensen}) and (\ref{eq:noiselevel_con}))} \\
\geq &1 - \exp{\frac{-2 \tau^4}{KL \|\z\|_\infty^4}}.
\end{split}
\end{equation}

Then Plug in the expected noise level derived in (\ref{eq:exp_noise}),
\begin{equation*}
\begin{split}
\P \{ \| \xf - \optx \|_2      \leq \cTwo(\delta) ( \sqrt{\frac{L}{m}}\|\z\|_2 + \frac{\tau}{ \sqrt{K}}  )    \} \\
 \geq 1 - \exp{\frac{-2 \tau^4}{KL \|\z\|_\infty^4}}.
\end{split}
\end{equation*}
and replacing $\tau / \sqrt{K}$ with $\tau$, the quantity on the right hand side of the equation then becomes $1 - \exp{\frac{-2 K \tau^4 }{L \|\z\|_\infty^4}}$ and
 we prove the theorem.

\subsection{Proof of Theorem~\ref{th:noisy_recon_ours}}
Now we consider the case that the BNSP is only satisfied for order $\sparse$
whereas there is no $\sparse-$sparse assumption on the true solution. Therefore, the algorithm can only guarantee the correctness of the $\sparse-$row sparse part 
and our best hope is to recover the best $\sparse-$row sparse approximation of the true solution. Let $\x_0$ be the best $\sparse-$row sparse approximation of the true solution $\optx$ and $\e$ denote the difference $\e = \optx - \x_0$. We rewrite the measurements to include the $\sparse-$row sparse approximation error as part of noise:
for $\ci = 1, 2, ..., K$,
\begin{equation}
\begin{split}
\yIci & = \AIci \optx + \zIci
= \AIci ( \x_0  + (\optx - \x_0 )) + \zIci \\
&  = \AIci \x_0  + \widetilde{\z}_\ci, 
\end{split}
\end{equation}
where $\widetilde{\z}_\ci  =  \AIci (\optx - \x_0 )+ \zIci =  \AIci \e + \zIci $.

To bound the distance of solution of $\ours$: $\Xsharp$ to the true solution $\optX$, we use the exactly $\sparse$ row sparse matrix $\Xtmp= ( \xtmp, \xtmp,..., \xtmp)$ as the bridge.
Since $\e = \optx - \xtmp$, we have:
$ \optX - \Xtmp  = ( \e , \e , ..., \e) $ and hence $ \|\Xtmp - \optX  \|_{2,2} = \sqrt{K} \| \e \|_2 $. Then the distance of $\Xsharp$ to the true solution $\optX$ can be decomposed into two parts:
\begin{equation}
\label{eq:decomp}
\begin{split}
 &  \| \Xsharp - \optX \|_{2,2} = \| \Xsharp - \Xtmp   + \Xtmp - \optX \|_{2,2}   \\ 
 & \leq \|\Xsharp- \Xtmp   \|_{2,2} + \|\Xtmp - \optX  \|_{2,2}\\
&= \|\Xsharp- \Xtmp   \|_{2,2} + \sqrt{K} \| \e \|_2  . 
\end{split}
\end{equation}

To bound the first term in (\ref{eq:decomp}): $\|\Xsharp- \Xtmp   \|_{2,2}$, we will use the recovery guarantee from the row sparse recovery result in Theorem~\ref{th:noisy_recon}~, which gives a upper bound of this term associated with the power of the noise matrix $\widetilde{\Z} = (\widetilde{\z}_1 ,\widetilde{\z}_2 ,...,\widetilde{\z}_K)$:
\begin{equation}
\begin{split}
\label{eq:z_tilde_form}
\| \widetilde{\Z} \|^2_{2,2} & = \sum_{\ci=1}^K \|\widetilde{\z}_\ci\|_2^2 = \sum_{\ci=1}^K \| \AIci \e + \zIci \|_2^2 \\
& = \sum_{\ci=1}^K  \sum_{\ri \in \I_\ci}  ( \langle \a \text{\footnotesize{$[i]$}} ,\e \rangle +  z \text{\footnotesize{$[i]$}} ) ^2.
\end{split}
\end{equation}
Then if we let  $\widetilde{\Xi} =  ( \langle \a\text{\footnotesize{$[i]$}}, \e \rangle  + z \text{\footnotesize{$[i]$}})^2 $
with $\a \text{\footnotesize{$[i]$}} , \z \text{\footnotesize{$[i]$}} $ generated uniformly from all rows of $\A$ and $\z$.
Since $\widetilde{\Xi}$ is non-negative, $\Xi \geq 0 $, the lower bound is $0$. 
The upper bound is derived using H{\"o}lders inequality:
\begin{equation}
\label{eq:eq_sep_bound}
\begin{split}
\widetilde{\Xi} & =  ( \langle \a \text{\footnotesize{$[i]$}} , \e \rangle + z \text{\footnotesize{$[i]$}} )^2 \leq  ( \|   \a \text{\footnotesize{$[i]$}} \cdot \e \|_1 +  \| \z\|_\infty )^2\\
 \leq &  ( \| \a \text{\footnotesize{$[i]$}} ^T \|_1 \|\e\|_\infty  + \| \z\|_\infty)^2\\
  \leq & ( \max_\ri \|  \a \text{\footnotesize{$[i]$}} ^T \|_1 \|\e\|_\infty  + \| \z\|_\infty)^2\\
  = & (\| \A \|_{\infty, 1} \|\e\|_\infty  + \| \z\|_\infty)^2 ,
\end{split}
\end{equation}
where $  \|\A \|_{\infty, 1} = \max_{ \ri \in [m]} \| \a \text{\footnotesize{$[i]$}}^T \|_1 $. Since $\A$ is deterministic with all bounded entries, $\|\A \|_{\infty, 1}$ is bounded.

From (\ref{eq:z_tilde_form}),  the expectation of $\| \widetilde{\Z} \|^2_{2,2} $ is
\begin{equation}
\label{eq:expectation2}
\begin{split}
\E & \| \widetilde{\Z} \|^2_{2,2}   =\sum_{\ci=1}^K   \sum_{\ri \in \I_\ci}  \E ( \langle \a[{\ri}] , \e \rangle )^2 + 2 \E z[\ri] \langle \a[\ri],  \e \rangle   \\
+ & \E z[{\ri}]^2 
= \frac{KL}{m} \| \A \e + \z \|_2^2.
\end{split}
\end{equation}

Obtaining the the lower and upper bound of $\widetilde{\Xi}$, we can apply Hoeffdings' inequality to get the tail bound of $\| \widetilde{\Z} \|_{2,2}^2$, which can be written as, for any $\tau > 0$,
\begin{equation}
\label{eq:tail}
\begin{split}
& \P \{\| \widetilde{\Z} \|^2_{2,2} -  \E\| \widetilde{\Z} \|^2_{2,2}  - \tau \leq 0 \} \\
& \geq 1  - \exp \frac{-2\tau^2}{KL (   \|\A \|_{\infty, 1} \|\e\|_\infty  + \| \z\|_\infty)^4}\\
\end{split}
\end{equation}

Then similar 
to prove Theorem \ref{th:noisy_recon_ours_exact_sparse}, 
here we consider the distance from the recovered solution $\Xsharp$ to the exactly $\sparse-$row sparse solution $\Xtmp$.
Let $\widetilde{\diffX} $ be $\widetilde{\diffX}= \frac{1}{\cTwo(\delta)}\| \Xsharp - \Xtmp \|_{2,2}$ and according to Theorem \ref{th:noisy_recon}, we have
\begin{equation}
\label{eq:cons}
\P \{ \| \widetilde{\diffX} - \| \widetilde{\Z} \|_{2,2} \leq 0 \} = 1. 
\end{equation}
Combing (\ref{eq:tail}) and (\ref{eq:cons}), we are able to conclude
\begin{equation*}
\label{eq:jobs_noiselevel_gen}
\begin{split}
 & \P   \{ \widetilde{\diffX}^2 -  \E \|\widetilde{\Z} \|^2_{2,2} - \tau^2 \leq 0 \}   \\
   & =\P\{  \widetilde{\diffX}^2 - \|\Z\|^2_{2,2} + \| \widetilde{\Z} \|^2_{2,2} - \E \|\widetilde{\Z}\|^2_{2,2}  - \tau^2 \leq 0  \}  \\
  & \geq \P\{  \widetilde{\diffX}^2 - \|\Z\|^2_{2,2} \leq 0 , \| \widetilde{\Z} \|^2_{2,2} - \E \|\widetilde{\Z}\|^2_{2,2}  - \tau^2 \leq 0  \}  \\
  & = \P\{  \widetilde{\diffX}^2 - \|\Z\|^2_{2,2} \leq 0 \} \P\{  \| \widetilde{\Z} \|^2_{2,2} - \E \|\widetilde{\Z}\|^2_{2,2}  - \tau^2 \leq 0  \}  \\
  &   \geq 1 - \exp \frac{-2\tau^4}{KL ( \|\A \|_{\infty, 1} \|\e\|_\infty  + \| \z\|_\infty)^4}.
\end{split}
\end{equation*}

We bound the expected square root of noise power:
\begin{equation}
\label{eq:noiselevel_sr}
\begin{split}
\P& \{ \widetilde{\diffX} \leq (\E \|\widetilde{\Z} \|^2_{2,2})^{1/2} + \tau \} \quad (\mbox{by \ (\ref{eq:noiselevel})} ) \\
  & \geq \P \{ \widetilde{\diffX}^2  \leq \E \|\widetilde{\Z} \|^2_{2,2} + \tau^2  \}\\
 & \geq 1 - \exp \frac{-2\tau^4}{KL (  \|\A \|_{\infty, 1} \|\e\|_\infty  + \| \z\|_\infty)^4}.
\end{split}
\end{equation}

Then, the final JOBS estimates $\xf$ is
$\xf  = \frac{1}{K} \sum_{\ci = 1}^K \x^{\boldsymbol{\sharp}}_{\ci}$ and by (\ref{eq:jensen}), we have: 
\begin{equation}
\label{eq:jobs_delta}
\begin{split}
\| \xf - \optx \|_2 
 & \leq   
   \frac{1}{\sqrt{K}} \| \Xsharp - \optX\|_{2,2}\\ 
   \text{ (by   (\ref{eq:decomp}))} & \leq \frac{1}{\sqrt{K}}  \| \Xsharp - \Xtmp \|_{2,2} +  \| \e \|_2 =\frac{\cTwo(\delta) \widetilde{\diffX}  }{\sqrt{K}}  + \| \e \|_2  \\
\end{split}
\end{equation}
Combing the results from (\ref{eq:noiselevel_sr}), (\ref{eq:jobs_delta}), we have:
\begin{equation}
\begin{split}
\P & \{ \| \xf - \optx \|_2  \leq  \frac{\cTwo(\delta) ((\E \| \widetilde{\Z}\|^2_{2,2})^{1/2} + \tau)  }{\sqrt{K}}  + \| \e \|_2   \}\\
 \geq  & \P \{ \frac{\cTwo(\delta) \widetilde{\diffX}  }{\sqrt{K}}  + \| \e \|_2   \leq  \frac{\cTwo(\delta) ((\E \| \widetilde{\Z} \|^2_{2,2})^{1/2} + \tau ) }{\sqrt{K}}  + \| \e \|_2   \}\\
= &  \P \{  \widetilde{\diffX} \leq   (\E \| \widetilde{\Z}\|^2_{2,2})^{1/2} + \tau     \}\\
\geq&  1 - \exp \frac{-2k\tau^4}{(   \|\A \|_{\infty, 1} \|\e\|_\infty  + \| \z\|_\infty)^4} . 
\end{split}
\end{equation}

Then plug in the expected noise level derived in (\ref{eq:expectation2}),
\begin{equation}
\begin{split}
\P& \{ \| \xf - \optx \|_2     \\
 & \leq \cTwo(\delta) ( \sqrt{\frac{L}{m}} \| \A \e + \z \|_2
 + \frac{\tau}{ \sqrt{K}}  )  + \|\e\|_2 \} \\
& \geq 1 - \exp \frac{-2\tau^4}{KL (   \|\A \|_{\infty, 1} \|\e\|_\infty  + \| \z\|_\infty)^4}.
\end{split}
\end{equation}
and replacing $\tau$ with $\tau / \sqrt{K}$, the quantity on the right hand side of the equation then becomes $1 - \exp \frac{-2K\tau^4}{L (   \|\A \|_{\infty, 1} \|\e\|_\infty  + \| \z\|_\infty)^4} $ and
we prove the theorem.


\subsection{Proof of Theorem \ref{th:noisy_bagging_exact_sparse}: performance bound of bagging for exactly $\sparse$-sparse signal recovery}
\label{sec:bagging_s}

Let $\xB$ be the solution of the bagging scheme, and it is an average over individual solved problems $\x^{\boldsymbol{B}}_1, \x^{\boldsymbol{B}}_2,..., \x^{\boldsymbol{B}}_K$: $\xB = \frac{1}{K}\sum_{\ci = 1}^K  \x^{\boldsymbol{B}}_\ci$. we consider the distance to the true solution $\optx$ to each estimate separately. Here, the desired upper bound is the square root of the expected power of each noise vector: $ (\E \zEi\|_2^2)^{1/2} = \sqrt{\frac{L}{m}} \|\z\|_2 $, where $\I$ is a multi-set of size $L$ with each element randomly sampled from $\{1,2,...,m\}$.
For $\tau >0$, we consider:
\begin{equation*}
\begin{split}
 \P  &  \{   \| \xB- \optx \|_2  -
 \cTwo(\delta)  (   (\E \|\zEi \|_2^2)^{1/2} + \tau ) \leq 0 \}  \\
= & \P   \{   \| \xB- \optx \|_2  -
 \cTwo(\delta)    (  (  (\E \|\zEi \|_2^2)^{1/2} + \tau)^2 )^{1/2}   \leq 0 \}  \\
\geq & \P   \{   \| \xB- \optx \|_2  -
 \cTwo(\delta)    (  \E \|\zEi \|_2^2+ \tau^2 )^{1/2}  \leq 0 \}  \\
= & \P   \{   \| \xB- \optx \|^2_2  -
 \cTwo^2(\delta)   (  \E \|\zEi \|_2^2 + \tau^2)  \leq 0 \} \\
 \end{split}
 \end{equation*}
Consider using the average of errors for each estimate: $\frac{1}{K}\sum_{\ci = 1}^K  \| \x^{\boldsymbol{B}}_\ci- \optx \|^2_2$, we have
 \begin{equation*}
 \begin{split}
 & = \P \{ \| \xB- \optx \|^2_2
-\frac{1}{K}\sum_{\ci = 1}^K  \| \x^{\boldsymbol{B}}_\ci- \optx \|^2_2\\
&+ \frac{1}{K}\sum_{\ci = 1}^K  \| \x^{\boldsymbol{B}}_\ci- \optx \|^2_2
 -\cTwo^2(\delta)   (  \E \|\zEi \|_2^2 + \tau^2)   \leq 0 \} \\
&\geq \P \{   \| \xB- \optx \|^2_2  -   \frac{1}{K}\sum_{\ci = 1}^K  \| \x^{\boldsymbol{B}}_\ci- \optx \|^2_2 \leq 0 , \\
& \frac{1}{K}\sum_{\ci = 1}^K  \| \x^{\boldsymbol{B}}_\ci- \optx \|^2_2  -\cTwo^2(\delta)   ( \E \|\zEi \|_2^2 + \tau^2)   \leq 0 \}
\end{split}
\end{equation*}
 By independence, we can factorize the two probabilities:
 \begin{equation*}
 \begin{split}
&= \P \{   \| \xB- \optx \|^2_2  -   \frac{1}{K}\sum_{\ci = 1}^K  \| \x^{\boldsymbol{B}}_\ci- \optx \|^2_2 \leq 0 \}  \\
&\cdot \P \{  \sum_{\ci = 1}^K  \| \x^{\boldsymbol{B}}_\ci - \optx\|^2_2-
K \cTwo^2(\delta)   ( \E \|\zEi \|_2^2 + \tau^2)  \leq 0 \}
\end{split}
\end{equation*}
By Jensens' Inequality, the first term is $1$ and
\begin{equation}
\label{eq:bagging_reduced}
\begin{split}
\P  &  \{   \| \xB- \optx \|_2  -
 \cTwo(\delta)  (   (\E \|\zEi \|_2^2)^{1/2} + \tau) \leq 0 \}  \\
  \geq& \P \{  \sum_{\ci = 1}^K  \| \x^{\boldsymbol{B}}_\ci - \optx\|^2_2-
K \cTwo^2(\delta)   ( \E \|\zEi \|_2^2 + \tau^2)  \leq 0 \} .
\end{split}
\end{equation}
From this procedure, we reduce the error bound for the bagging algorithm to bound the sum of individual errors.

We let random variable $\underline{\x} = \| \x \text{\scriptsize $(\I)$} - \optx \|^2_2$, where $\x \text{\scriptsize $(\I)$}$ is the solution from $\ell_1$ minimization on bootstrap samples of size $L$: $\x \text{\scriptsize $(\I)$}= \argmin \| \x\|_1 \st \| \yEi - \AEi \|_2^2 \leq \epsilon $, where $\I$ denotes a bootstrap sample. 
All $\underline{\xci } = \| \x^{\boldsymbol{B}}_\ci - \optx \|^2_2$ are realizations generated i.i.d. from the distribution of $\underline{\x}$. We proceed the proof using the following lemma that gives
the tail bound of the sum of i.i.d. bounded random variables, and its proof
follows a similar procedure as proving Hoeffdings' inequality (details in Appendix~\ref{sec:proof_lemma_sum_iid}). 

\begin{lemma}[Tail bound of the sum of i.i.d. bounded Random variables]
\label{lemma:sum_of_rvs}
Let $Y_1, Y_2,..., Y_n$ be i.i.d. observations of bounded random variable $Y$: $a \leq Y \leq b$ and the expectation $\E Y$ exists, for any $\epsilon > 0$, then
\begin{equation}
\label{eq:sum_bd_tail}
\P \{ \sum_{i = 1}^n Y_i \geq n \epsilon \} \leq \exp\{ - \frac{2 n ( \epsilon - \E Y) ^2}{(b - a)^2} \} .
\end{equation}
\end{lemma}

In this case, we consider the lower bound $a$ and the upper bound $b$ of random variable $\underline{\x}$. Clearly $\underline{\x} \geq 0$, we therefore set $a =0$. The upper bound is obtained from the error bound of $\ell_1$ minimization in Theorem~\ref{th:noisy_recon_l1}.
For all $\I$: 
\begin{equation}
\label{eq:smv_bd}
\P \{ \| \x \text{\scriptsize $(\I)$} - \optx \|^2_2 - \cTwo^2(\delta) \| \zEi\|^2_2 \leq 0 \} =1,  
\end{equation}
According to the norm equivalence inequality
\begin{equation}
\label{eq:norm_ieq}
  \| \zEi \|^2_2  \leq  ( \sqrt{L}\| \zEi \|_\infty) ^2 \leq ( \sqrt{L}\| \z\|_\infty) ^2 =  L\| \z\|^2_\infty.
\end{equation}
and we set $b =\cTwo^2(\delta) L \| \z\|^2_\infty  $.

Now we can apply use (\ref{eq:sum_bd_tail}) to analyze our problem. By (\ref{eq:bagging_reduced}), the $\epsilon$ in (\ref{eq:sum_bd_tail}) is: $\epsilon =
\cTwo^2(\delta)   ( \E \|\zEi \|_2^2 + \tau^2)  $ , then 
\begin{equation}
\label{eq:sum_upbd}
  \P \{  \sum_{\ci = 1}^K  \| \xci - \optx\|^2_2-
k  \epsilon \geq 0 \} \leq \exp\{ - \frac{2 K ( \epsilon - \E \underline{\x} ) ^2}{\cTwo^4(\delta) L^2 \| \z\|^4_\infty  } \}.
\end{equation}
To simplify the right hand side, we consider:
$   \E \underline{\x} = \E \| \x - \optx \|_2^2 =  \frac{1}{|m^L|}\sum_{\I}\| \xSolEi - \optx \|_2^2 $. From our bound in (\ref{eq:smv_bd}), it implies that
\begin{equation*}
\begin{split}
\P & \{ \frac{1}{|m^L|}\sum_{\I}\| \x_{\I} - \optx \|_2^2 \leq \frac{1}{|m^L|}\sum_{\I} \cTwo^2(\delta)\| \z_\I \|_2^2 \}  = 1,
\end{split}
\end{equation*}
which is equivalent to
\begin{equation}
\label{eq:boundex}
\begin{split}
 \E \| \xSolEi - \optx \|_2^2 &  \leq \frac{1}{|m^L|}\sum_{\I} \cTwo^2(\delta)\| \zEi \|_2^2 \\
    & =  \E \ \cTwo^2(\delta)\| \z_{\I} \|_2^2   =   \cTwo^2(\delta) \E \| \z_{\I} \|_2^2 . 
\end{split}
\end{equation}
Then we have
\begin{equation}
\begin{split}
& \epsilon - \E \underline{\x}
 =  \cTwo^2(\delta)   ( \E \|\zEi \|_2^2 + \tau^2)   - \E \| \x - \optx \|_2^2\\
 \geq   & \cTwo^2(\delta)   ( \E \|\zEi \|_2^2 + \tau^2) - \cTwo^2(\delta) \E \| \z_{\I} \|_2^2  = \cTwo^2(\delta) \tau^2. \\
\end{split}
\end{equation}
The right hand side of (\ref{eq:sum_upbd}) is upper bounded by $\exp\{ - \frac{2 K  \tau^4}{ L^2 \| \z\|^4_\infty  } \}.$

\subsection{Proof of Theorem \ref{th:noisy_bagging_dm} performance bound of bagging for general sparse signal recovery}
%
%

In this section, we are working with the case when the true solution $\optx$ is a general sparse signal, which sparsity level may exceed $\sparse$ and the $\sparse-$sparse approximation error is no longer necessarily zero.
Let $\epsilon_\sparse$ denote the sparse approximation error $\epsilon_\sparse = \cOne(\delta) \sparse^{-1/2} \| \e\|_{1}$, we consider the following:
\begin{equation*}
\begin{split}
 \P  &  \{   \| \xB- \optx \|_2  - (\epsilon_\sparse+ 
\cTwo (\delta) (  \sqrt{ \frac{L}{m}}  \| \z \|_2 + \tau)) \leq 0 \}\\
 =   & \P \{  \| \xB- \optx \|_2^2 - 
 ( 
 \epsilon_\sparse
 + \cTwo(\delta)  (  \sqrt{ \frac{L}{m}}  \| \z \|_2  
+ \tau))^2 \leq 0 \}  \\
  \geq   & \P \{   \| \xB- \optx \|_2^2 - 
 ( ( 
 \epsilon_\sparse
 + \cTwo(\delta)  \sqrt{ \frac{L}{m}}  \| \z \|_2)^2  
+  \cTwo^2(\delta) \tau^2 ) \leq 0 \}  \\
 \end{split}
\end{equation*}
We let $\epsilon'= ( 
 \epsilon_\sparse+\cTwo(\delta)   \sqrt{ \frac{L}{m}}  \| \z \|_2)^2  +  \cTwo^2(\delta) \tau^2 )$ and we consider using the averages of the errors $ \frac{1}{K}\sum_{\ci = 1}^K  \| \x^{\boldsymbol{B}}_\ci- \optx \|^2_2 $ as an intermediate term. Repeat the same proving technique as we did in (\ref{eq:bagging_reduced}), we have
\begin{equation*}
 \P \{   \| \xB- \optx \|_2^2 - \epsilon'\} \geq  \P \{  \sum_{\ci = 1}^K  \| \x_\ci^{\boldsymbol{B}} - \optx\|^2_2-
K \epsilon'  \leq 0 \}.
\end{equation*}


According to Lemma~\ref{lemma:sum_of_rvs}~, we have:
\begin{equation}
\label{eq:sum_upbd_general}
\begin{split}
  \P \{  \sum_{\ci = 1}^K  \| \x_\ci^{\boldsymbol{B}} - \optx\|^2_2-
K \epsilon' \geq 0 \} \leq \exp\{ - \frac{2 K ( \epsilon' - \E \underline{\x}) ^2}{ (b' - a'   )^2} \}.
\end{split}
\end{equation}
Here $a' = 0$, and $b' = ( 
\epsilon_\sparse + {\cTwo(\delta)} \sqrt{L}\| \z \|_\infty)^{2} $. The lower bound $a'$ is set to zero since $\underline{\x}$ is non negative and the upper bound $b'$ is obtained using Theorem~\ref{th:noisy_recon_l1} and plug in the upper bound of the noise power as derived in (\ref{eq:norm_ieq}).


We consider the term $\epsilon'  - \E \underline{\x} = (\cOne(\delta) \sparse^{-1/2} \| \e\|_{1} + \cTwo(\delta)   \sqrt{ \frac{L}{m}}  \| \z \|_2)^2  +  \cTwo^2(\delta) \tau^2  - \E \| \x - \optx \|_2^2$. We upper bound the expected value of $\underline{\x}$ in the same approach as in (\ref{eq:boundex}). From Theorem~\ref{th:noisy_recon_l1}, for all $\I$:
\begin{equation*}
\P \{ \| \x \text{\scriptsize $(\I)$} - \optx\|_2^2 \leq
(\epsilon_\sparse + \cTwo (\delta)\| \zEi \|_2 ) ^2 \}  = 1.
\end{equation*}
Therefore,
\begin{equation}
\label{eq:boundex_g}
\begin{split}
\P & \{ \E \| \xEi - \optx \|_2^2 \leq  \E (  \epsilon_\sparse  
+ \cTwo (\delta)\| \zEi \|_2 ) ^2 \} = 1.
\end{split}
\end{equation}

Because $f(x) = x^2 $ is a convex function, and therefore by Jensens' inequality, we have:
\begin{equation*}
 (\E \|\zEi \|_2) ^2 \leq  \E \|\zEi  \|^2_2.
\end{equation*}
Because square root $x^{1/2}$ is a increasing function with $x$, therefore taking square root preserves the sign of inequality:
\begin{equation}
\label{eq:bd_convex}
 \E \|\zEi  \|_2 \leq  ( \E \|\zEi  \|^2_2)^{1/2}.
\end{equation}

Then from (\ref{eq:boundex_g}), we have:
\begin{equation*}
\begin{split}
  \E & \| \xEi - \optx \|_2^2  \leq \E (  \epsilon_\sparse  
+ \cTwo (\delta)\| \zEi \|_2 ) ^2\\
 & = \epsilon_\sparse^2  + \cTwo^2 (\delta) \E \|\zEi \|^2_2 
+ 2 \epsilon_\sparse  \cTwo (\delta) \E \| \zEi \|_2  \quad    \\
 &(\mbox{by (\ref{eq:bd_convex}}))\\
& \leq  \epsilon_\sparse^2 + \cTwo^2 (\delta) \E \| \zEi \|^2_2 + 2 \epsilon_\sparse \cTwo (\delta) ( \E \|\z_\I \|^2_2)^{1/2}  \\
& = (  \epsilon_\sparse + \cTwo (\delta) (\E \|\zEi \|^2_2)^{1/2} )^2 
\end{split}
\end{equation*}

From previous result, we have  $( \E \|\zEi  \|^2_2)^{1/2} = \sqrt{\frac{L}{m}} \| \z \|_2$ and therefore $\epsilon' = (\epsilon_\sparse+\cTwo(\delta)    (\E \|\zEi \|^2_2)^{1/2})^2  +  \cTwo^2(\delta) \tau^2 $.
Then we can bound the term $\epsilon' - \E \| \x - \optx \|_2^2$:
\begin{equation}
\begin{split}
& \epsilon' - \E \| \x - \optx \|_2^2 \\
= & (\epsilon_\sparse+\cTwo(\delta)    (\E \|\zEi \|^2_2)^{1/2})^2  +  \cTwo^2(\delta) \tau^2  -  \E \| \x - \optx \|_2^2\\
 \geq & ((\epsilon_\sparse+\cTwo(\delta)    (\E \|\zEi \|^2_2)^{1/2})^2 +  \cTwo^2(\delta) \tau^2 \\
& - (\epsilon_\sparse 
+ \cTwo(\delta)    (\E \|\zEi \|^2_2)^{1/2} )^2
  =\cTwo(\delta)^2 \tau^2. \\
\end{split}
\end{equation}
The bound in (\ref{eq:sum_upbd_general}) can be upper bounded by
\begin{equation*}
\begin{split}
  \P \{  \sum_{\ci = 1}^K  \| \x_\ci^{\boldsymbol{B}}- \optx\|^2_2-
K  \epsilon' \geq 0 \} \leq \exp\{ - \frac{2 K \cTwo^4(\delta) \tau^4}{ (b'    )^2} \}. 
\end{split}
\end{equation*}
where $b' = (\cOne(\delta) \sparse^{-1/2} \| \e\|_{1} + {\cTwo(\delta)} \sqrt{L}\| \z \|_\infty)^{2} $.

\subsection{Sufficient condition: Theorem~\ref{cor:suff} from Sample Complexity for gaussian and bernoulli random matrices}


  We would like to connect the BRIP constant of $\AJ$ to RIP constants all submatrices $\AIci$s. First, we consider using $V$ to represent the number of distinct measurements of bootstrapping samples of size $L$, and $ 0 \leq L \leq m$. 
 Pick $d < L$ being the smallest number of distinct samples that we would like to have hold with probability at least $1 - \alpha$, which is \begin{equation}
\P (V \geq d) \geq 1 - \alpha.
\end{equation}
The relationship of $d , L, m $ and the $\alpha$
given the rest variables can be found in Appendix~\ref{app:bp} in 
(\ref{eq:distinct_alpha}).


Let $V_1, V_2, ..., V_k$ count the number of distinct measurements of all sub-measurements $\yIcOne, \yIcTwo,..., \yIcK$.
Because of the bootstrap procedure, $V_i$s are i.i.d. distributed as random variable $V$. Consider the probability that all the $V_i \geq d$, we have:
\begin{equation}
\label{eq:prob_all}
\begin{split}
 \P \{ \forall \ci \ V_\ci \geq d \}  = \P  \{ \bigcap_{\ci = 1}^K  \{ V_\ci \geq d \} \}
 &  =  \prod_{\ci = 1}^K \P \{ V_\ci \geq d   \}\\
 & \geq (1 - \alpha)^K .\\
\end{split}
\end{equation}
We would like to calculate the BRIP constant of $\AJ = \diag ( \A{[\I_1]} , \A{[\I_2]}, ..., \A{[\I_K]} )$. 
To simplify the process, we first consider the same certainty level $\mu_{J}$ and lower bound of number of distinct samples $d$ for each
for the standard RIP constant for each sub-marix $\AIci$. Entries of the distinct rows of each sub-matrix come from Gaussian distribution.
 According to Theorem \ref{th:rdmtx},  if we have enough distinct measurements   $d \geq \Const \delta^{-2}(2 \sparse \ln(n/ 2 \sparse) + \ln(\mu_J^{-1}))$, then for $\mu_J , d$, all $\ci = 1, 2,...,K$
\begin{equation}
\label{eq:conf_bound}
\P \{   \delta_{2 \sparse}(\AIci ) \leq  \delta | V_\ci \geq d \} \geq 1 - \mu_J,
\end{equation}
Note that, here the RIP constant of $\AIci$ considers the RIP constant on distinct rows of $\AIci$.

Now we consider the BRIP constant of $\AJ$ , given the condition that all sub-matrices has at least $d$ distinct measurements.
According to (\ref{eq:oursRIP}) in Proposition \ref{prop:factorization}, we have
\begin{equation*}
\begin{split}
&  \P \{ \delta_{2s|\B}(\AJ) = \max_{ \ci  } \delta_{2 \sparse}(\AIci ) \leq  \delta
| \forall \ci \ V_\ci \geq d  \} & \\
& = \P \{  \forall \ \ci = 1, 2, ..., m :  \delta_{2 \sparse}(\AIci ) \leq  \delta
| V_\ci \geq d \}    & \\
&  = 1 -   \P \{  \exists \ \ci = 1,2 ,..., m :  \delta_{2 \sparse}(\AIci ) >  \delta
| V_\ci \geq d \}  &
\end{split}
\end{equation*}

Note that although $\AIci$ are not mutually independent, we can employ union bound:
\begin{equation*}
\begin{split}
&  \geq 1 -   \sum_{i = 1}^K \P \{   \delta_{2 \sparse }(\AIci ) >  \delta
| V_\ci \geq d \}    & \\
&  \geq  1 - K                    \mu_{J}.
\end{split}
\end{equation*}
Finally, we consider the BRIP constant of $\AJ$

\begin{equation}
\label{eq:bound}
\begin{split}
&\P \{  \delta_{2s|\B}(\AJ) \leq \delta \} = \P \{ \max_{ \ci } \delta_{2 \sparse }(\AIci ) \leq  \delta \}\\
&= \P \{ \max_{ \ci  } \delta_{2 \sparse }(\AIci ) \leq  \delta| \forall \ci \ V_\ci \geq d \}  \P \{  
\forall \ci  \  V_\ci \geq d  \}\\
&+ \P \{ \max_{ \ci  } \delta_{2 \sparse }(\AIci ) \leq  \delta| \exists \ci \ V_\ci < d \}  \P \{  \exists \  V_\ci < d \}\\
& \geq    \P \{ \max_{ \ci  } \delta_{2 \sparse }(\AIci ) \leq  \delta| \forall \ci \ V_\ci \geq d \}  \P \{  
\forall \ci  \  V_\ci \geq d  \}  \\
 \end{split}
 \end{equation}
 We here drop the second term to get a lower bound. According to (\ref{eq:prob_all}): $\P  \{ \exists \ V_i < d \} \leq 1 - (1 - \alpha)^K$, which is fairly small when $\alpha$ is small. The choice of $\alpha$ is preferred to be small in practice and the bound is a good for practical proposes since it is tighter when $\alpha$ is smaller. 
 Then by (\ref{eq:prob_all}), we have:
\begin{equation}
\label{eq:bound}
\begin{split}
&\P \{ \max_{ \ci  } \delta_{2 \sparse }(\AIci ) \leq  \delta \}\\
&\geq \P \{ \max_{ \ci  } \delta_{2\sparse}(\AIci ) \leq  \delta| \forall \ci \ V_\ci \geq d \}  \P \{  
\forall \ci  \  V_\ci \geq d  \}\\
& \geq   ( 1 - K \mu_{J}) (1 - \alpha)^K.
 \end{split}
 \end{equation}

To simplify the bound in (\ref{eq:bound}), we would like to achieve
\begin{equation}
\label{eq:wewant}
 ( 1 - K \mu_{J}) (1 - \alpha)^K \geq 1 - \mu, \mbox{for some } \mu \in [0, 1].
\end{equation}
Namely,
\begin{equation}
\begin{split}
&( 1 - K \mu_{J}) (1 - \alpha)^K \geq 1 - \mu \\
\Longleftrightarrow &( 1 - K \mu_{J})  \geq  \frac{1 - \mu} {(1 - \alpha)^K} \\
\Longleftrightarrow & \mu_{J}   \leq  \frac{1}{K} - \frac{1 - \mu} {K(1 - \alpha)^K} = \frac{(1 - \alpha)^K - (1 - \mu)}{K(1 - \alpha)^K}.
\end{split}
\end{equation}

 According to Theorem~\ref{th:rdmtx}, (\ref{eq:wewant}) can be achieved if
\begin{equation}
\label{eq:sample_comp_jobs}
\begin{split}
 d & > \Const \delta^{-2} (2 \sparse \ln(n/ 2 \sparse) + \ln(\mu_J^{-1}) ) \\
 &\geq \Const \delta^{-2} (2\sparse \ln(n/ 2 \sparse) - \ln( \frac{(1 - \alpha)^K - (1 - \mu)}{K(1 - \alpha)^K} )\\
& =   \Const \delta^{-2} (2 \sparse \ln(n/ 2 \sparse) + \ln K  +\ln( \frac{(1 - \alpha)^K}{(1 - \alpha)^K - (1 - \mu)} )).
\end{split}
\end{equation}

Replace $\delta_{2 \sparse}( \mu_{J} , d( \alpha, m , L) )$ by its upper bound $\sqrt{2}-1$
and therefore the sample complexity is $ \O (2 \sparse \ln(n/ 2 \sparse) + \ln K+\ln( \frac{(1 - \alpha)^K}{(1 - \alpha)^K - (1 - \mu)} ) )$ and the constant depends on $\mu$ and $\alpha$. Both the last two terms of (\ref{eq:sample_comp_jobs}) : $\ln K  +\ln( \frac{(1 - \alpha)^K}{(1 - \alpha)^K - (1 - \mu)})$ are introduced by the uncertainty introduced by the bootstrap procedure and they are all non-decreasing with respect to $K$. This theorem also matches the RIP condition that increasing $K$ by adding extra multi-sets $\I$s, the RIP constant will guarantee to be non-decreasing.

Note that there are some limitations of this theorem. The prove follows standard RIP condition for sparse recovery, however, the range that RIP condition guarantees are not wide enough: in this case, the worst case performance is limited by the worst $\AIci$. As a result the probably of success being guaranteed in (\ref{eq:bound}) has $(1-\alpha)^K$, which will vanish fast if $K$ is large and then the bound becomes quite loose. Also, there is an implicit condition while proving: to guarantee all the probabilities to be between zero and one, while $K > 1$, equation (\ref{eq:wewant}) implicitly implies that $\mu \geq \alpha$. This means that, the certainty of the performance of the algorithm is limited by the certainty level of the minimum number of distinct measurements across each column in the measurement matrix. This implicit assumption makes sense however it is a bit conservative to estimate performances in practice. 

\section{Simulations }  
\label{sec:simulation}

In this section, we perform sparse recovery on simulated data to study the performance of our algorithm. In our experiment, all entries of $ \A \in \R^{m \times n}$ are i.i.d. samples from the standard normal distribution $\mathcal{N} (0, 1)$. This simulation setting is the same as the one in the analysis part by multiplying a normalization factor $\frac{1}{m}$. The signal dimension $ n = 200 $ and various numbers of measurements from $50$ to $2000$ are explored. 
For the ground truth signals, their sparsity levels are $\sparse=50$, and the non-zeros entries are sampled from the standard gaussian with their locations being generated uniformly at random. 
For the noise processes $\z$, which entries are sampled i.i.d. from $\mathcal{N} ( 0 , \sigma^2) $, with variance $\sigma^2 = 10^{-\SNR/10} \| \A \x \|_2^2$, where $\SNR$ represents the Signal to Noise Ratio.
In our experiment, we study three different ratios: $\SNR=0, 1$ and $2$ dB.

We use the ADMM implementation of Block (Group) LASSO~\cite{admm} to solve the unconstraint form of $\ours$ 
, in which the parameter $\lambda^{(k,L)}$ balances the least squares fit and the joint sparsity penalty:
\vspace{-0.05in}
\begin{equation}
\label{eq:bmmv_uo}
\min_{\X}  \ \lambda^{(K,L)} \| \X \|_{1,2} + \frac{1}{2} \sum_{\ci =1}^K \| \yIci - \AIci \xci  \|_2^2.
\end{equation}
The same solver is used to solve $\ell_1$ minimization with $K=1$ for 
a fair comparison with all other algorithms.

We study how the number of estimates $K$ as well as the bootstrapping ratio $L/m$ affects the result. In our experiment, we take $K = 30, 50, 100$, while the bootstrap ratio $L/m $ varies from $0.1$ to $1$. 
We report the Signal to Noise Ratio (SNR) as the error measure for recovery : $ \SNR = 10 \log_{10} \| \x - \optx \|_2^2 / \| \optx \|_2^2$ averaged over $20$ independent trials. For all algorithms, we evaluate $\lambda^{(K,L)}$ at different values from $.01$ to $200$ and then select optimal values that gives the maximum averaged SNR over all trials.

\begin{figure}[!h]
\centering
\parbox{0.27\textwidth}{\centering{JOBS}}
\parbox{0.205\textwidth}{\centering{Bagging}}
\hfil
\hfil
\vspace{-0.2cm}
\captionsetup[subfigure]{labelformat=empty}
\subfloat[]
{\raisebox{\dimexpr 3cm-\height}[0pt][0pt]{\rotatebox[origin=c]{90}{$m = 50$}}}
\subfloat[]
{\label{fg:l11}\includegraphics[trim = {0 0 0 20}, clip, width=0.25\textwidth]{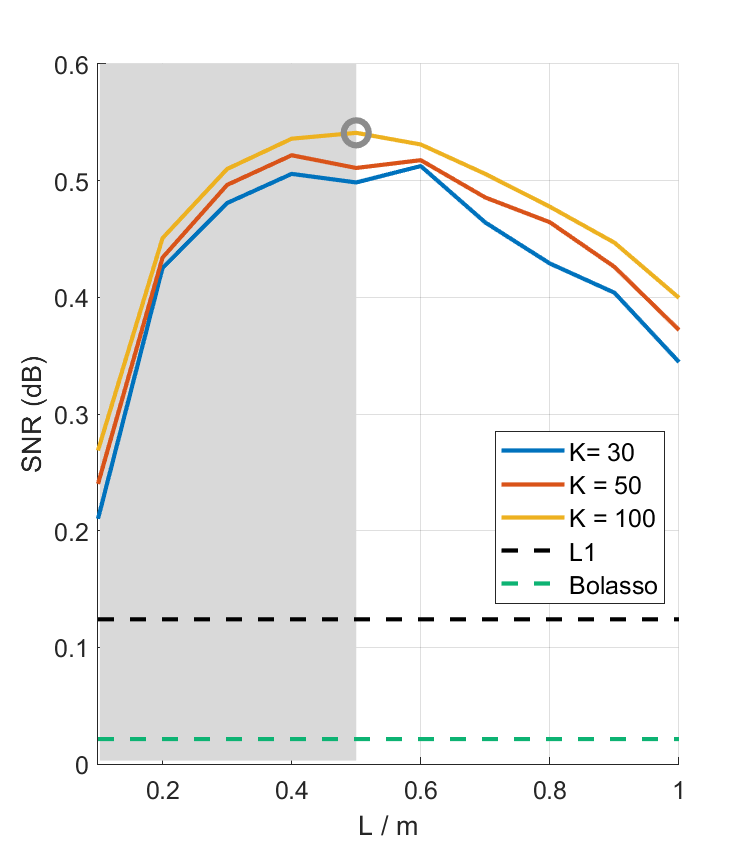}}
\subfloat[]
{\label{fg:l12}\includegraphics[trim = {0 0 0 20}, clip,
width=0.25\textwidth]{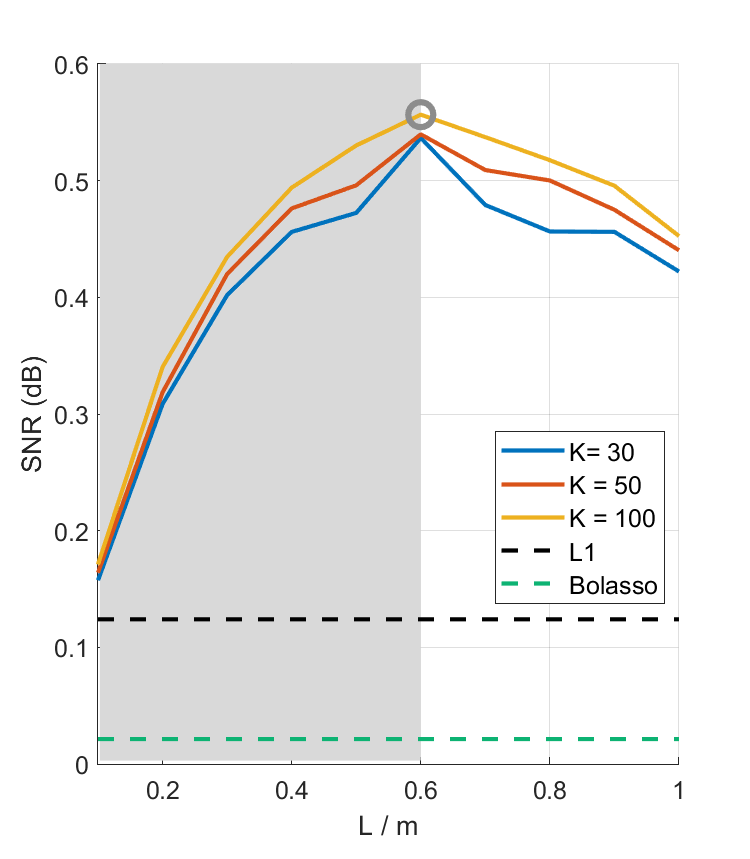}}\\
\vspace{-0.5cm}
\subfloat[]
{\raisebox{\dimexpr 3cm-\height}[0pt][0pt]{\rotatebox[origin=c]{90}{$m = 75$}}}
\subfloat[Ours][]
{\label{fg:l11}\includegraphics[trim = {0 0 0 20}, clip, width=0.25\textwidth]{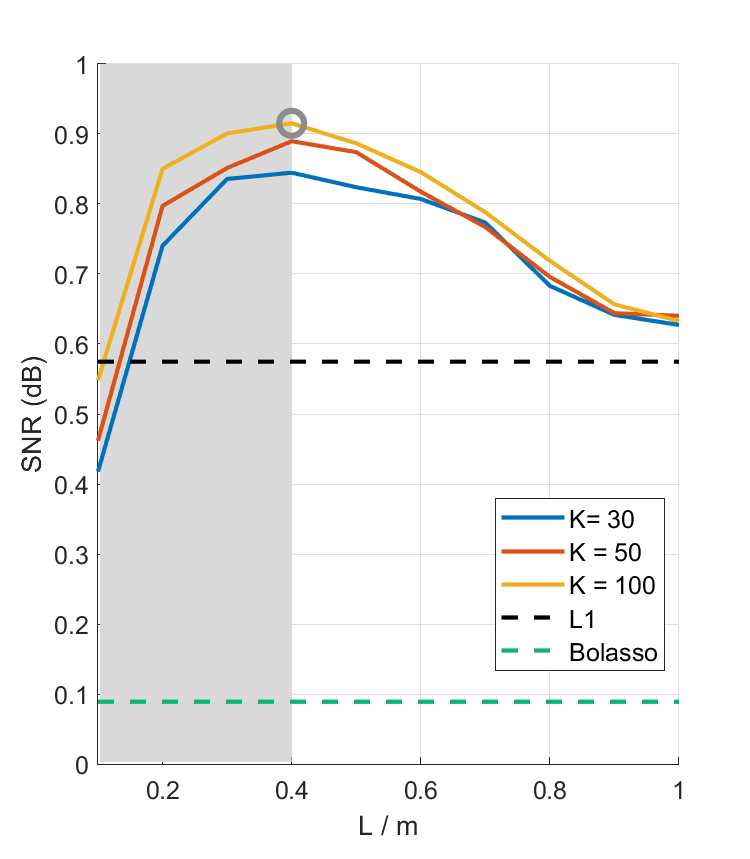}}
\subfloat[Bagging][]
{\label{fg:l12}\includegraphics[trim = {0 0 0 20}, clip, width=0.25\textwidth]{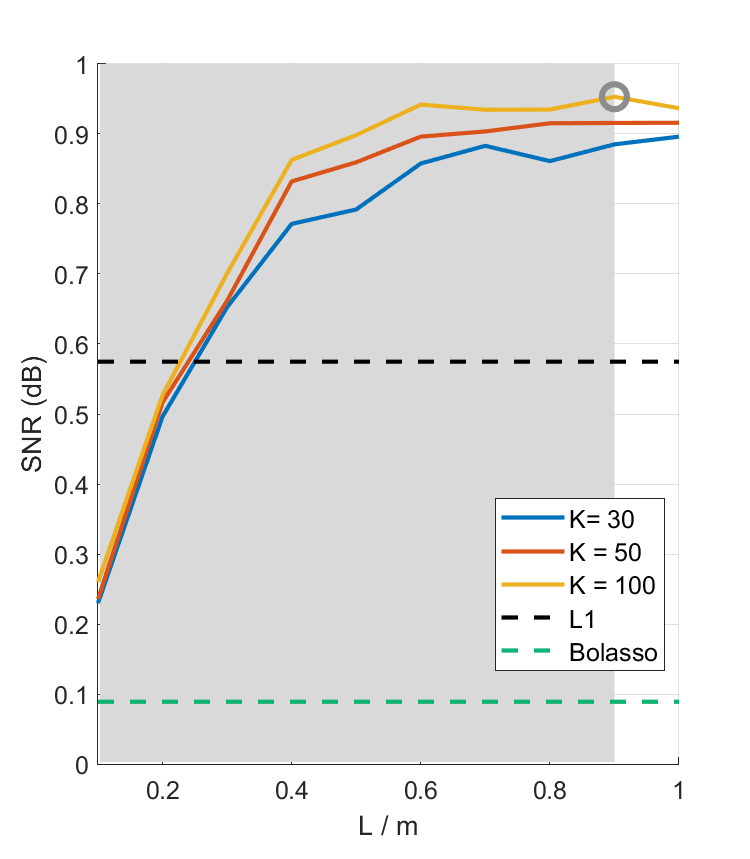}}\\
\vspace{-0.5cm}
\subfloat[]
{\raisebox{\dimexpr 3cm-\height}[0pt][0pt]{\rotatebox[origin=c]{90}{$m = 100$}}}
\subfloat[Ours][]
{\label{fg:l11}\includegraphics[trim = {0 0 0 20}, clip, width=0.25\textwidth]{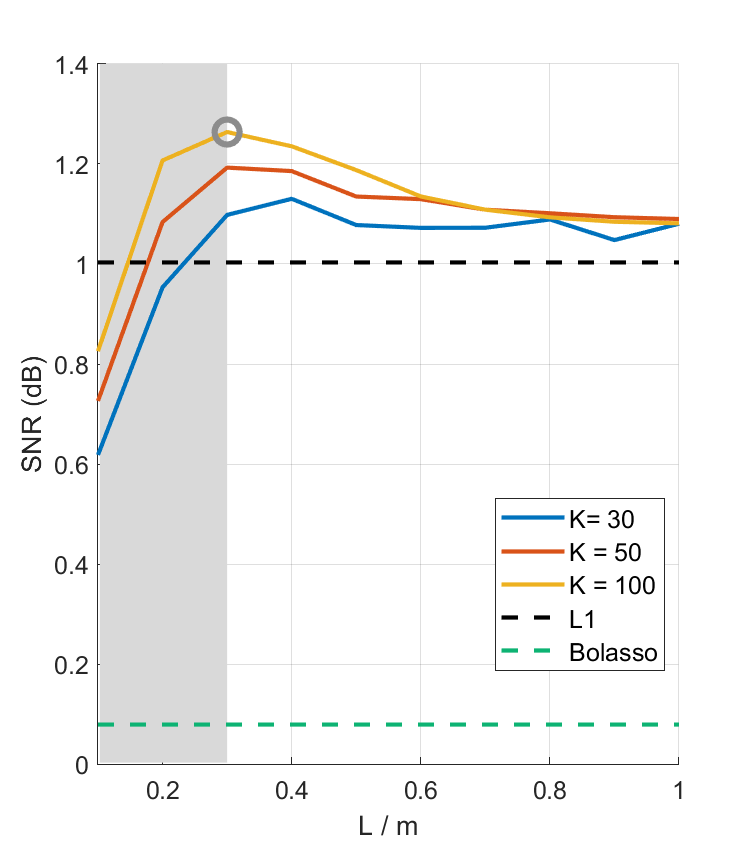}}
\subfloat[Bagging][]
{\label{fg:l12}\includegraphics[trim = {0 0 0 20}, clip, width=0.25\textwidth]{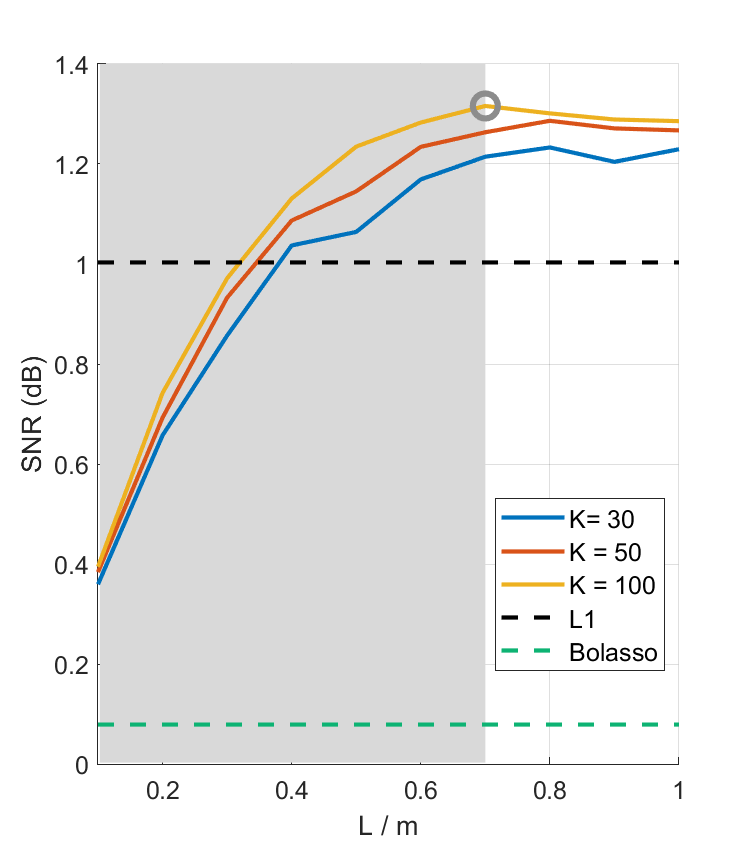}}\\
%
\vspace{-0.5cm}
\subfloat[]
{\raisebox{\dimexpr 3cm-\height}[0pt][0pt]{\rotatebox[origin=c]{90}{$m = 150$}}}
\subfloat[Ours][]
{\label{fg:l11}\includegraphics[trim = {0 0 0 20}, clip, width=0.25\textwidth]{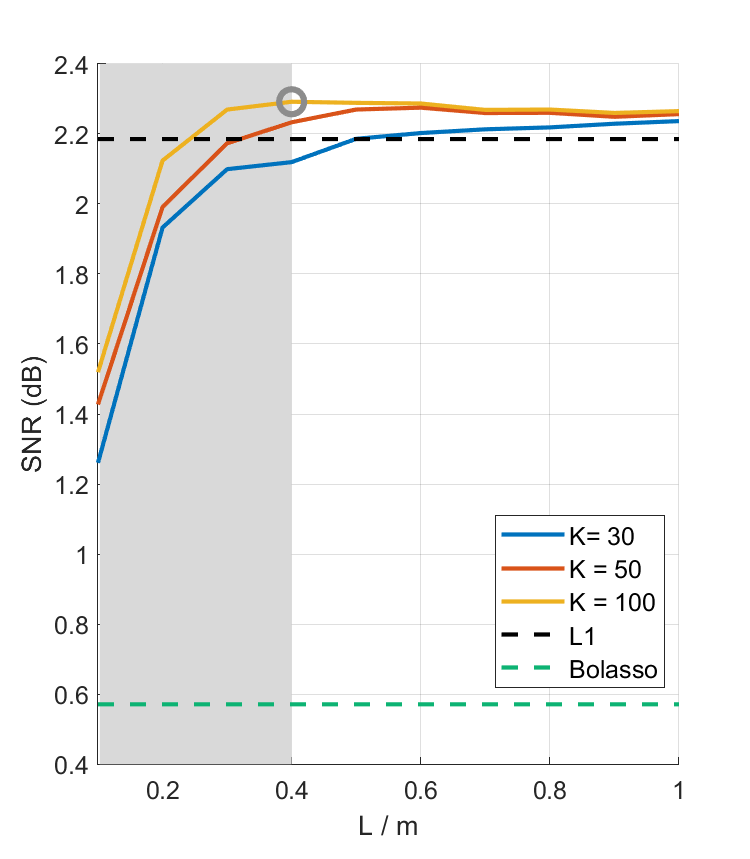}}
\subfloat[Bagging][]
{\label{fg:l12}\includegraphics[trim = {0 0 0 20}, clip, width=0.25\textwidth]{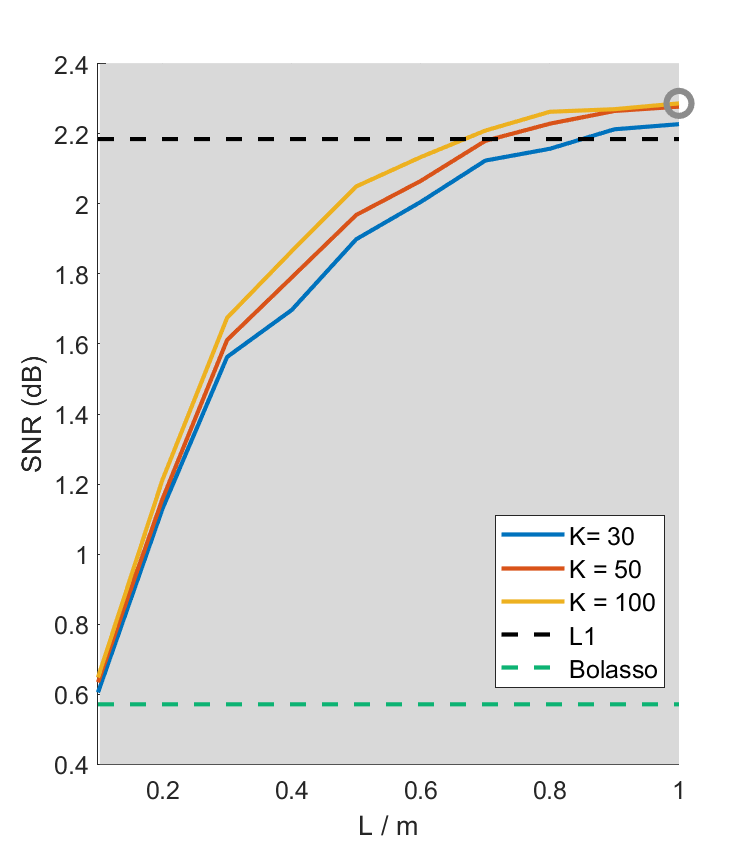}}
\vspace{-0.3cm}
\caption{Performance curves for JOBS, Bagging and Bolasso with different $L,K$ as well as $\ell_1$ minimization. $\SNR=0 $ dB and
the number of measurements $m = 50 , 75, 100, 150$ from top to bottom.
The grey circle highlights the peaks of JOBS, Bagging and Bolasso and the grey area highlights the bootstrap ratio at the peak point. JOBS requires smaller $L/m$ than Bagging to achieve peak performance. JOBS and Bagging outperform $\ell_1$ minimization and Bolasso when $m$ is small. }
\vspace{-0.2cm}
\label{fg:exp1}
\end{figure}

\begin{figure}[!h]
\centering
\hfil
\parbox{0.27\textwidth}{\centering{JOBS}}
\parbox{0.205\textwidth}{\centering{Bagging}}
\hfil
\vspace{-0.2cm}
\captionsetup[subfigure]{labelformat=empty}
\subfloat[]
{\raisebox{\dimexpr 3cm-\height}[0pt][0pt]{\rotatebox[origin=c]{90}{$m = 200$}}}
\subfloat[Ours][]
{\label{fg:l11}\includegraphics[trim = {0 0 0 20}, clip, width=0.25\textwidth]{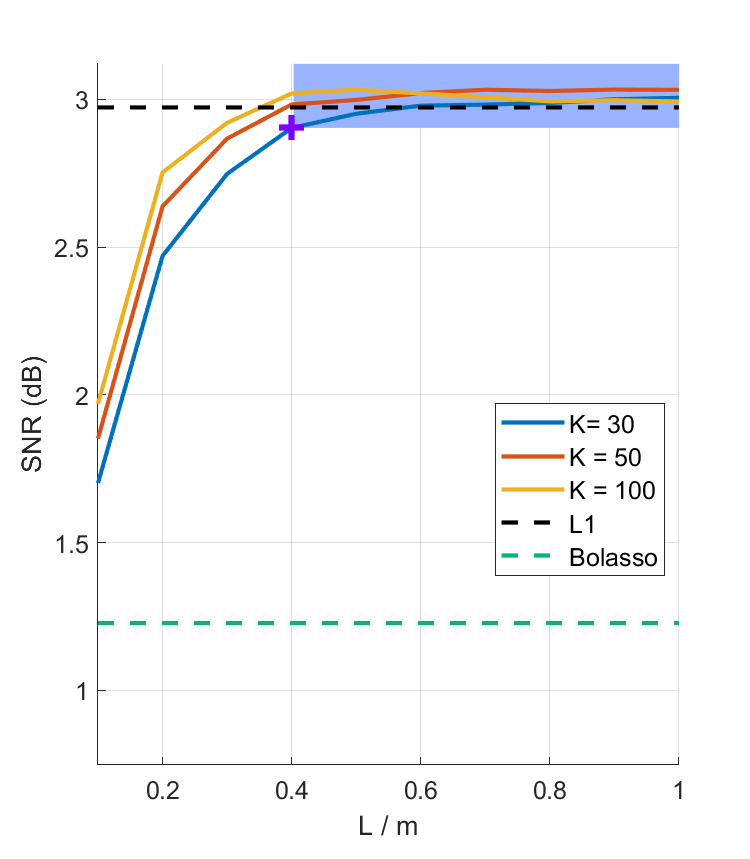}}
\subfloat[Bagging][]
{\label{fg:l12}\includegraphics[trim = {0 0 0 20}, clip, width=0.25\textwidth]{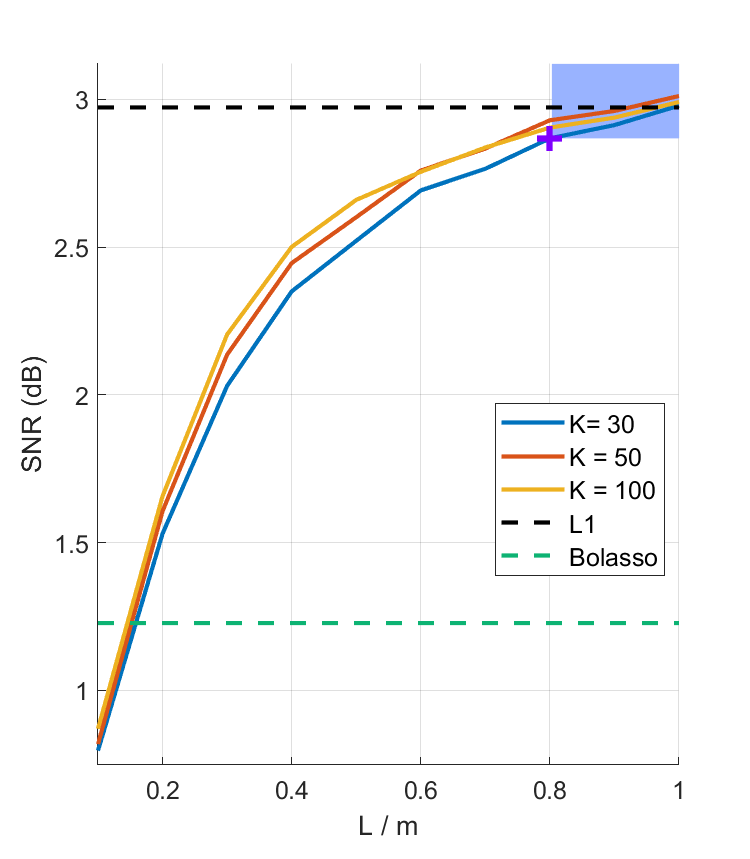}}\\
\vspace{-0.5cm}
\subfloat[]
{\raisebox{\dimexpr 3cm-\height}[0pt][0pt]{\rotatebox[origin=c]{90}{$m = 500$}}}
\subfloat[Ours][]
{\label{fg:l11}\includegraphics[trim = {0 0 0 20}, clip, width=0.25\textwidth]{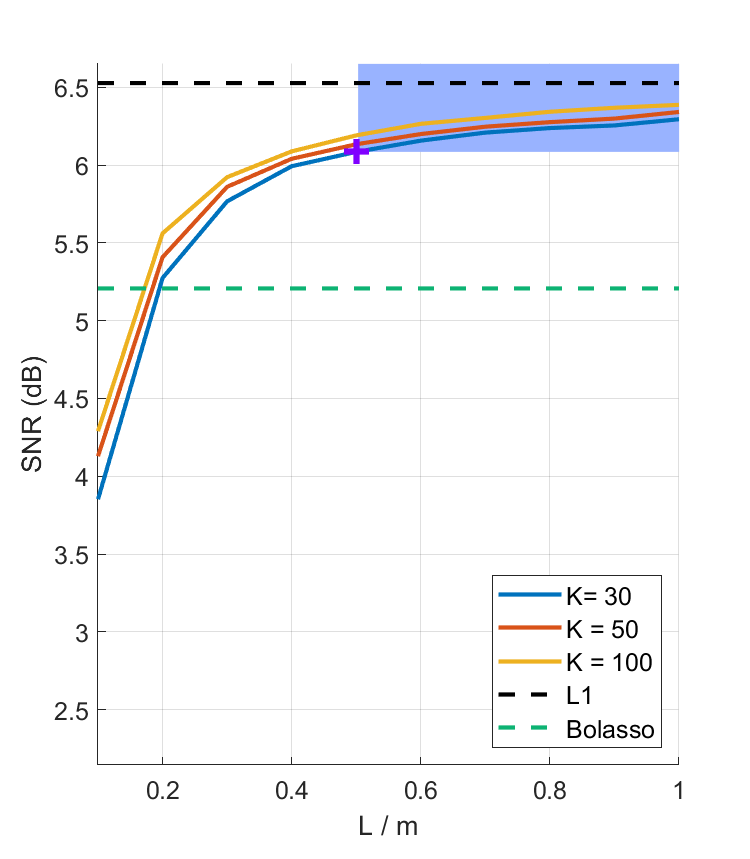}}
\subfloat[Bagging][]
{\label{fg:l12}\includegraphics[trim = {0 0 0 20}, clip, width=0.25\textwidth]{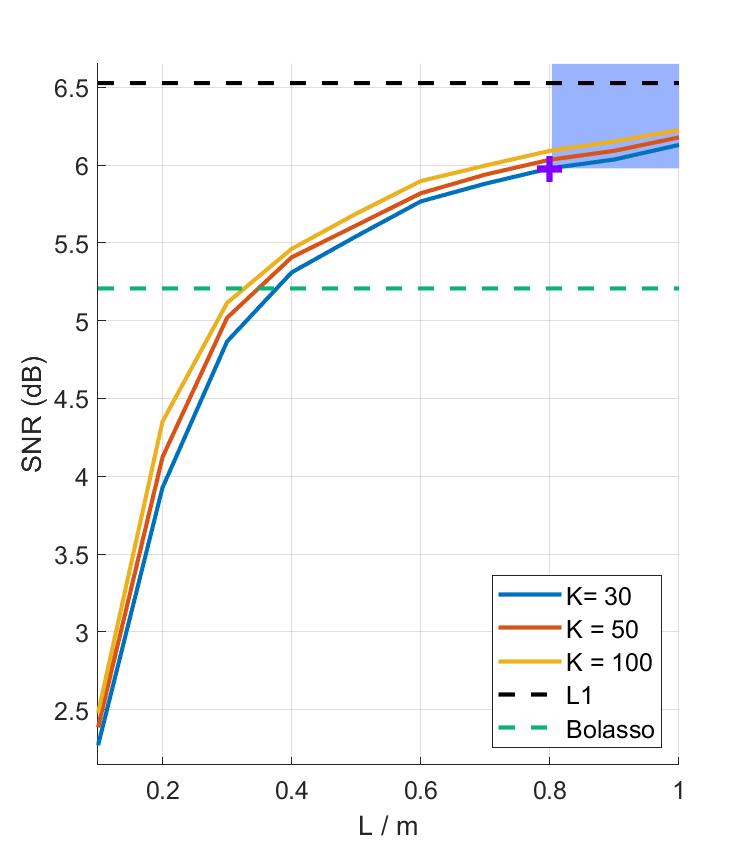}}\\
%
\vspace{-0.5cm}
\subfloat[]
{\raisebox{\dimexpr 3cm-\height}[0pt][0pt]{\rotatebox[origin=c]{90}{$m = 2000$}}}
\subfloat[Ours][]
{\label{fg:l11}\includegraphics[trim = {0 0 0 20}, clip, width=0.25\textwidth]{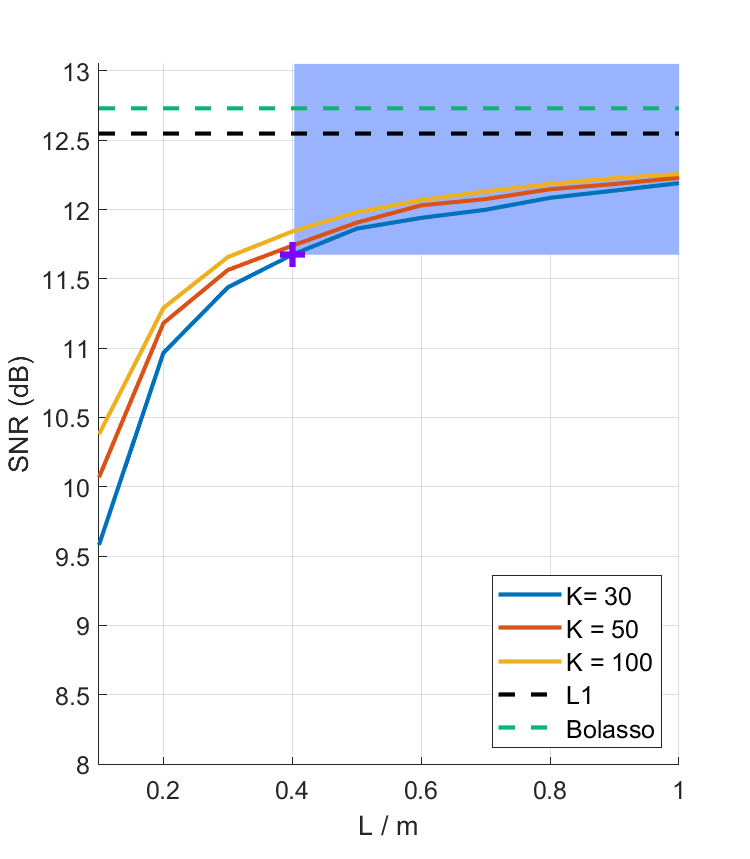}}
\subfloat[Bagging][]
{\label{fg:l12}\includegraphics[trim = {0 0 0 20}, clip, width=0.25\textwidth]{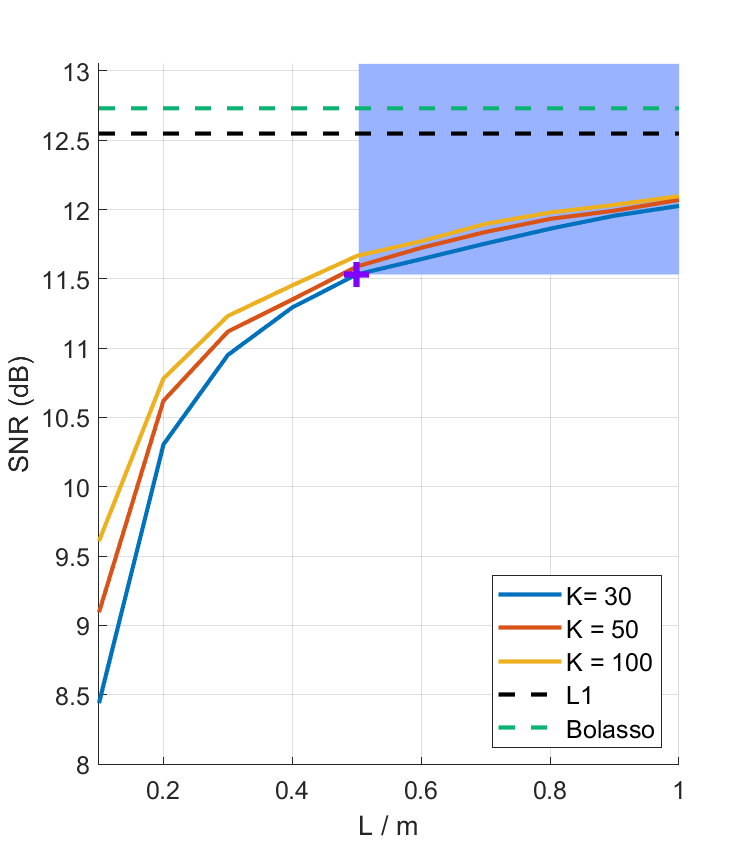}}
\vspace{-0.3cm}
\caption{Performance curves for JOBS, Bagging and Bolasso with different $L,K$ as well as $\ell_1$ minimization. $\SNR = 0 $ dB and the number of measurements $m = 200,500,2000$ from top to bottom. The purple highlighted area is where at least $95 \%$ of the best performance achieved and within which the minimum $K \times L $ is illustrated by purple cross.
JOBS requires smaller $L/m$ than Bagging to achieve acceptable performance. }
\label{fg:exp2}
\end{figure}

\begin{figure*}[!h]
\centering
\subfloat[][SNR = 0 dB]
{\label{fg:pv_snr0}\includegraphics[trim = {0 0 0 20}, clip, width=0.32\textwidth]{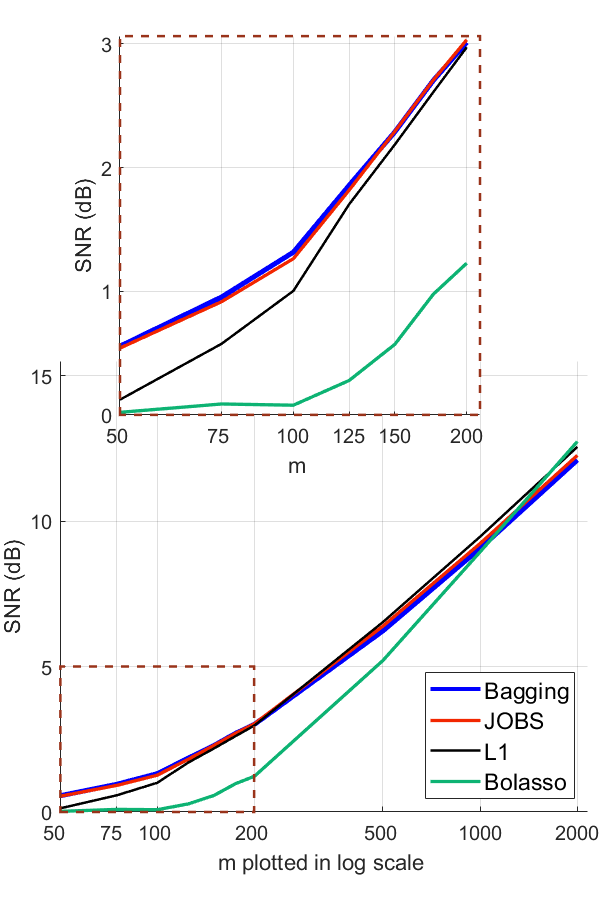}}
\hfill
\subfloat[][SNR = 1 dB]
{\label{fg:pv_snr1}\includegraphics[trim = {0 0 0 20}, clip, width=0.32\textwidth]{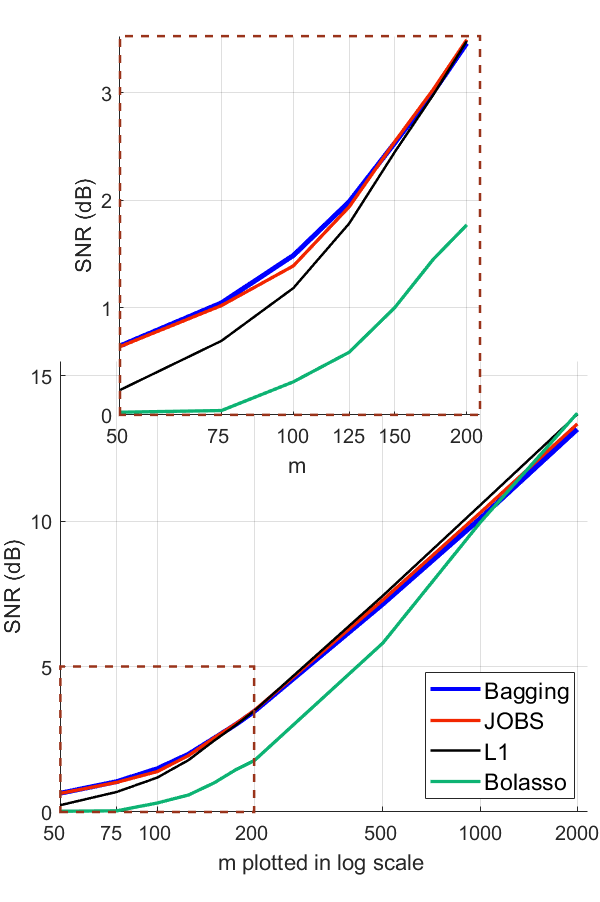}} 
\hfill
\subfloat[][SNR = 2 dB]
{\label{fg:pv_snr2}\includegraphics[trim = {0 0 0 20}, clip, width=0.32\textwidth]{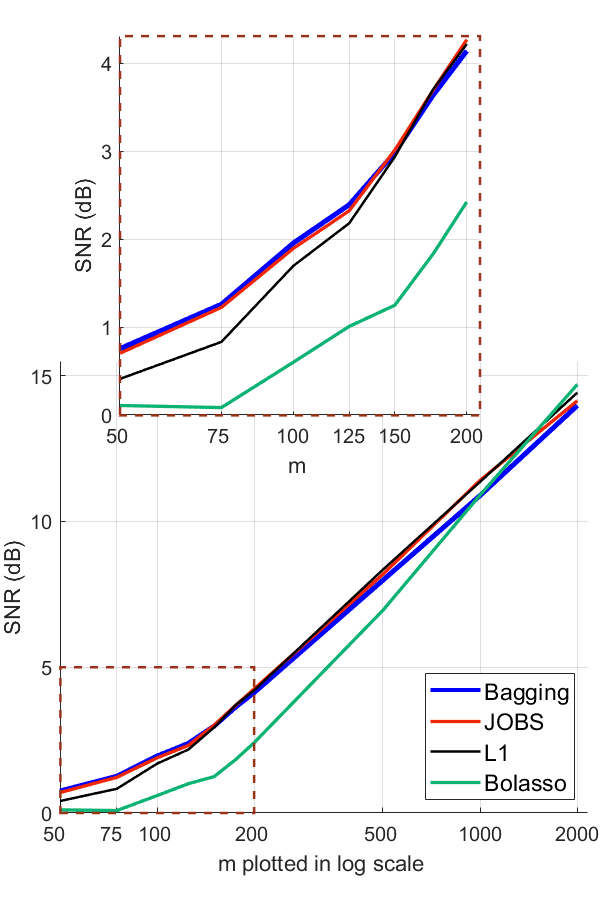}}
\caption{The best performance among all choices of $L$ and $K$ with various number of measurements for JOBS, Bagging, Bolasso and $\ell_1$ minimization. \hspace{\textwidth} While $m$ is small, and lower SNR, the margin between JOBS and $\ell_1$ minimization is larger (zoomed-in figures on the top row). }
\label{fig:peaks_wtP}
\end{figure*}

\subsection{Performance of JOBS, Bagging, Bolasso and $\ell_1$ minimization}
Beside JOBS, Bagging and Bolasso with the same parameters $K,L$ and $\ell_1$ minimization are studied.  The result are plotted in Figure~\ref{fg:exp1} and Figure~\ref{fg:exp2}. The colored curves shows the cases with various number of estimates $K$. The grey circle highlights the best performance and the grey area highlights the optimal bootstrap ratio $L/m$. In those figures, for each condition with a choice of $K, L$, the information available to JOBS, Bagging and Bolasso algorithms is identical, and $\ell_1$ minimization always access to all $m$ measurements.
We plot the performance of JOBS, Bagging with various $L/m$ ratio and $K$.  The performance of $\ell_1$ minimization is depicted with the black dashed lines, while the best Bolasso performance is plotted using light green dashed lines.

From Figure~\ref{fg:exp1}, we see that when $m$ is small, JOBS can outperform $\ell_1$ minimization. As $m$ decreases, the margin increases. 
It is rather surprising that with a low number of measurements ($m$  is between $\sparse$ to $3\sparse$: $50 - 150$),
and with very small $L$ and $K$ ($L$ about only $30\% - 40 \%$ of the entire set of measurements and $K$ at $30$), our algorithm is already quite robust and outperforms all other algorithms with the same $(K, L)$ parameters used. Although in terms of the best performance limit, JOBS and Bagging are similar, Bagging requires $L$ to be around $60 \% - 90 \%$ of the entire set of measurements to achieve comparable performance as JOBS.
The correct prior information with the row sparsity on multiple estimates may especially show its advantage while the amount of information is limited. However, when the level measurement is high enough, bootstrapping loses its advantages and $\ell_1$ becomes the preferred strategy. 

Figure~\ref{fg:exp2} shows the performance with on large number of measurements ($m = 200, 500, 2000$), revealing that JOBS requires a much smaller $L$ to a comparable performance to Bagging.
The purple highlighted area is where at least $95 \%$ of the best performance achieved and within which the minimum $K \times L $ is illustrated by purple cross.
The purple region are larger and the locations of purple crosses in various $m$ are much further left for our algorithm compared to the ones in Bagging. This implies that
although the local maxima for Bagging and JOBS are similar, much smaller $KL$ are required to obtain an acceptable performance.
The bootstrapping ratio $L/m$ is $40\%-50\%$ for JOBS, $50\%-80\%$ for Bagging and $K = 30$ for both algorithms to achieve at least $95\%$ of the best performance for each algorithm. The subsampling variation that we will illustrate in Section~\ref{sec:simulation2} and the result in Figure~\ref{fg:exp1_var_largeM} has an similar advantage.
This result is quite promising for large number of measurements. Especially in the streaming setting where utilizing all data at once in a batch algorithm like $\ell_1$ minimization is not applicable.
 When the process is stationary, employing our methods to enforce joint sparsity on multiple local windows boosts performances and the recovery is reasonable with a smaller amount of data. 

Figure~\ref{fg:pv_snr0},  Figure~\ref{fg:pv_snr1} and Figure~\ref{fg:pv_snr2} depict the best performance for various schemes: JOBS, Bagging, Bolasso and $\ell_1$ minimization with SNR values at 0, 1, 2 dB respectively.
For the first three algorithms, 
 the peak values are found among different choices of parameters $K$ and $L$ that we explored. We see that when the number of measurements $m$ is low, JOBS and Bagging outperform $\ell_1$ minimization.  The larger the noise level (lower SNR), the larger the margin. Although the performance limits of JOBS and Bagging are very similar,
Figure~\ref{fg:exp1} shows that JOBS achieves comparable performance to Bagging with significantly smaller $L , K $ values. JOBS and Bagging tend to converge to $\ell_1$ minimization as $m$ increases. Bolasso only performs similarly to other algorithms for a large $m$ and slightly outperforms all other algorithms when $m= 2000$. 
\vspace{-0.2cm}
\subsection{Subsampling Variation to Ensure Distinct Samples }  
\label{sec:simulation2}
Random sampling with replacement (bootstrapping) likely creates duplicates within the samples. Although it simplifies the analysis, in practice, duplicate information does not add much value. Therefore, in this simulation, we conduct a more practical variation of JOBS scheme. To ensure the distinctness within each sample, each time we conduct subsampling, instead of bootstrapping:
for each bootstrap sample $\I_{\ci}$, $L$ distinct samples are generated by random sampling without replacement from $m$ measurements. There are a few differences compared to the previous bootstrapping scheme:
{\it (i)} For each subset $\I_{\ci}$ , the information contained for this subsampling variation will be at least as much as the original scheme.
{\it (ii)} When the subsampling ratio $L/ m = 1$, both the subsampling variation of JOBS and Bagging coincides with MMV version of $\ell_1$ minimization, and therefore in this case, they all behave the same as $\ell_1$ minimization.
The original bootstrapping scheme, $L$ is required to be a much larger number than $m$ to observe all $m$ samples.

Similarly to the previous section, we study how the number of estimates $K$ as well as the subsampling ratio $L/m$ affects the result. The variation is also adopted in Bagging and Bolasso. The subsampling version of Bagging is the stochastic approximation of Subagging estimator (short for \textbf{Sub}sampling \textbf{Agg}regat\textbf{ing})~\cite{subagging,subaggingTwo}. Here, for simplicity of the terms, we refer all methods by their original names.  All the experimental settings are the same as the previous one except the bootstrapping resampling scheme is replaced by subsampling for each subset $\I_{\ci}$.

\begin{figure}[!h]
\centering
\parbox{0.27\textwidth}{\centering{JOBS}}
\parbox{0.205\textwidth}{\centering{Bagging}}
\hfil
\vspace{-0.2cm}
\captionsetup[subfigure]{labelformat=empty}
\subfloat[]
{\raisebox{\dimexpr 3cm-\height}[0pt][0pt]{\rotatebox[origin=c]{90}{$m = 50$}}}
\subfloat[]
{\label{fg:l11}\includegraphics[trim = {0 0 0 20}, clip, width=0.25\textwidth]{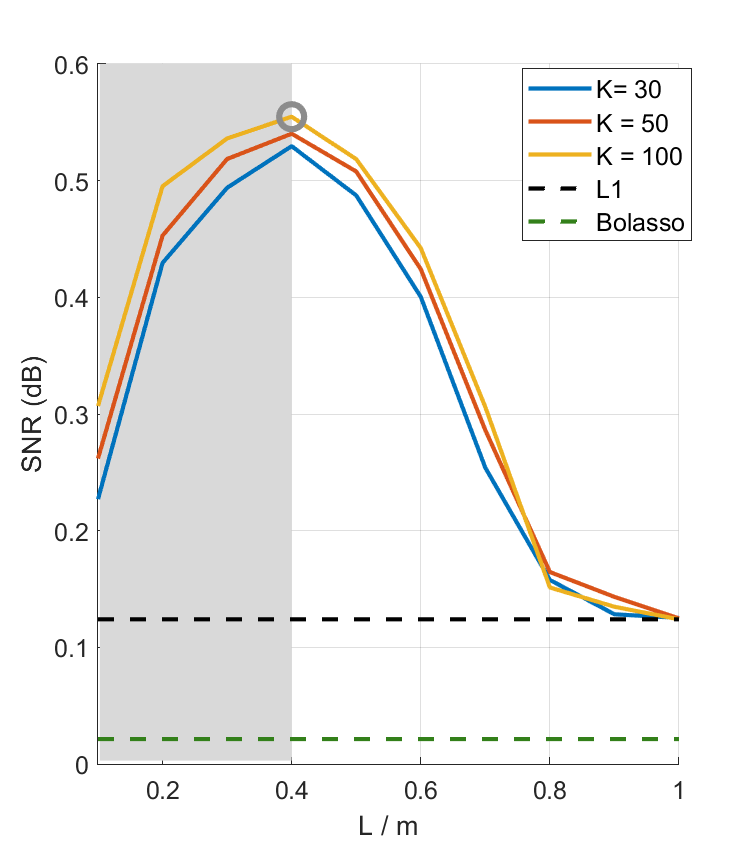}}
\subfloat[]
{\label{fg:l12}\includegraphics[trim = {0 0 0 20}, clip,
width=0.25\textwidth]{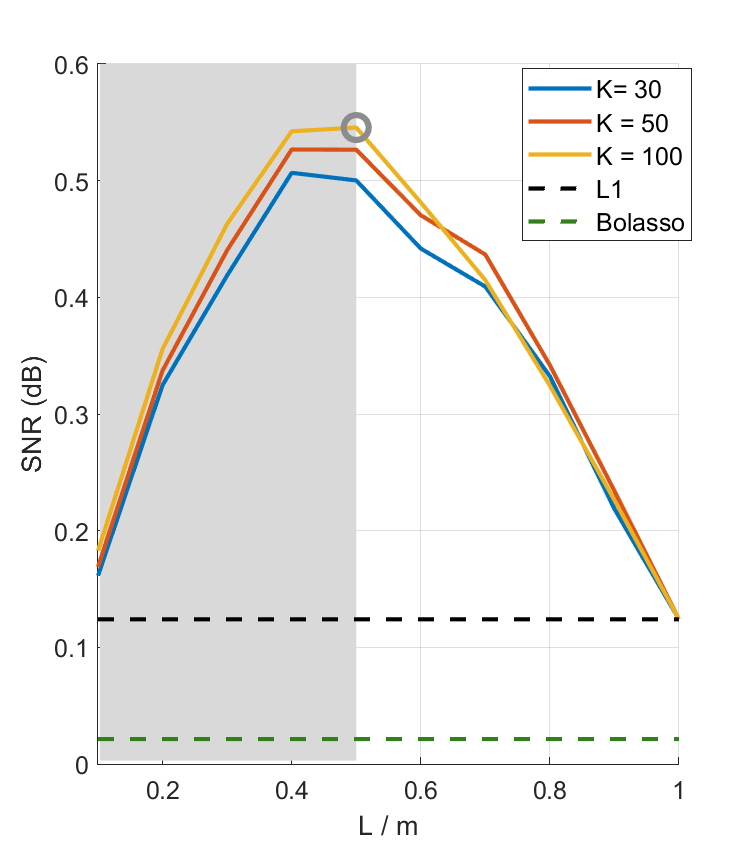}}\\
\vspace{-0.5cm}
\subfloat[]
{\raisebox{\dimexpr 3cm-\height}[0pt][0pt]{\rotatebox[origin=c]{90}{$m = 75$}}}
\subfloat[Ours][]
{\label{fg:l11}\includegraphics[trim = {0 0 0 20}, clip, width=0.25\textwidth]{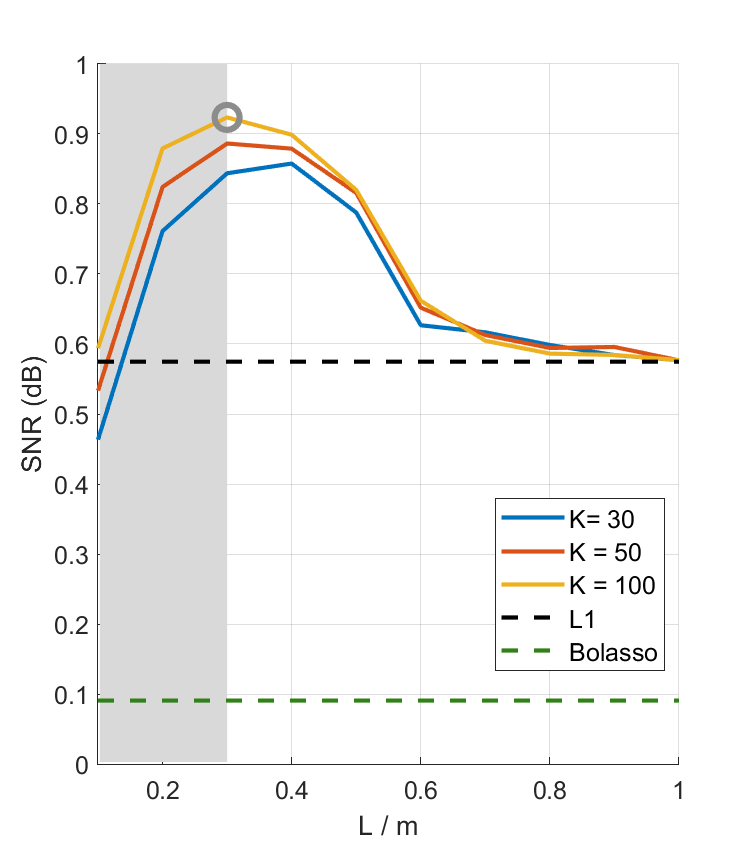}}
\subfloat[Bagging][]
{\label{fg:l12}\includegraphics[trim = {0 0 0 20}, clip, width=0.25\textwidth]{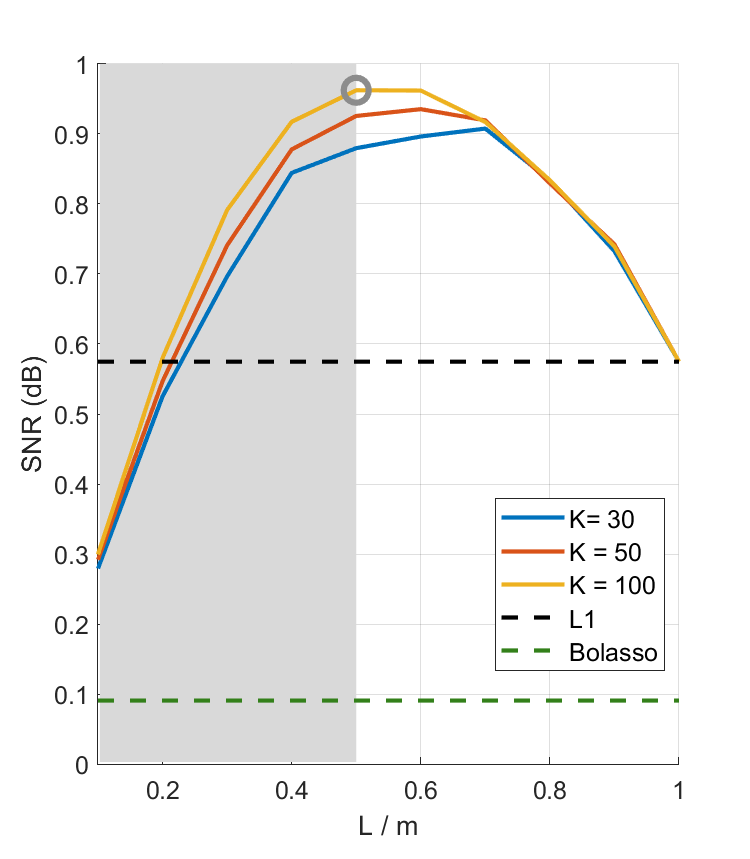}}\\
\vspace{-0.5cm}
\subfloat[]
{\raisebox{\dimexpr 3cm-\height}[0pt][0pt]{\rotatebox[origin=c]{90}{$m = 100$}}}
\subfloat[Ours][]
{\label{fg:l11}\includegraphics[trim = {0 0 0 20}, clip, width=0.25\textwidth]{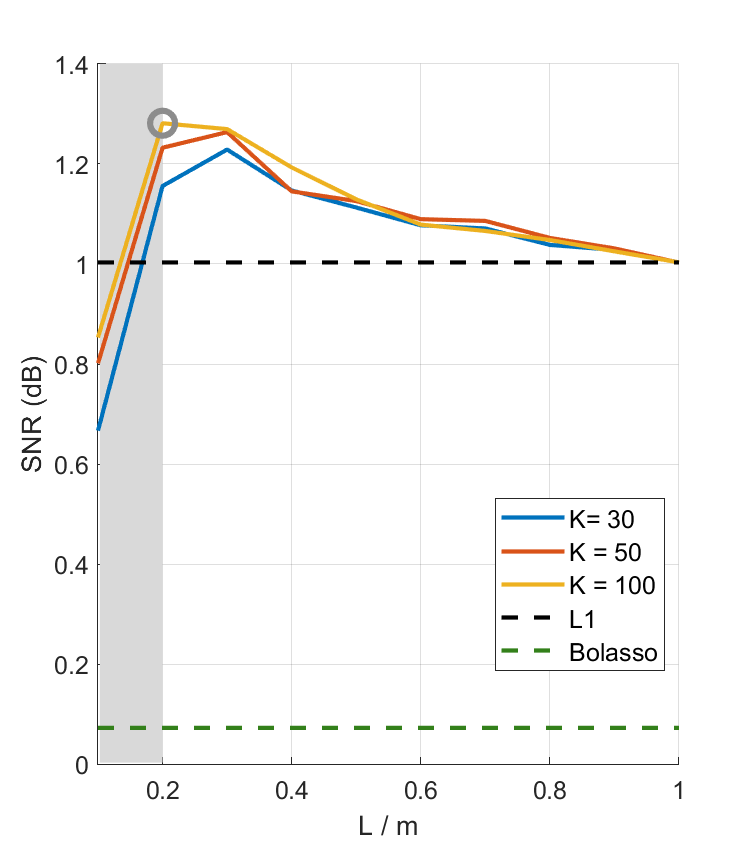}}
\subfloat[Bagging][]
{\label{fg:l12}\includegraphics[trim = {0 0 0 20}, clip, width=0.25\textwidth]{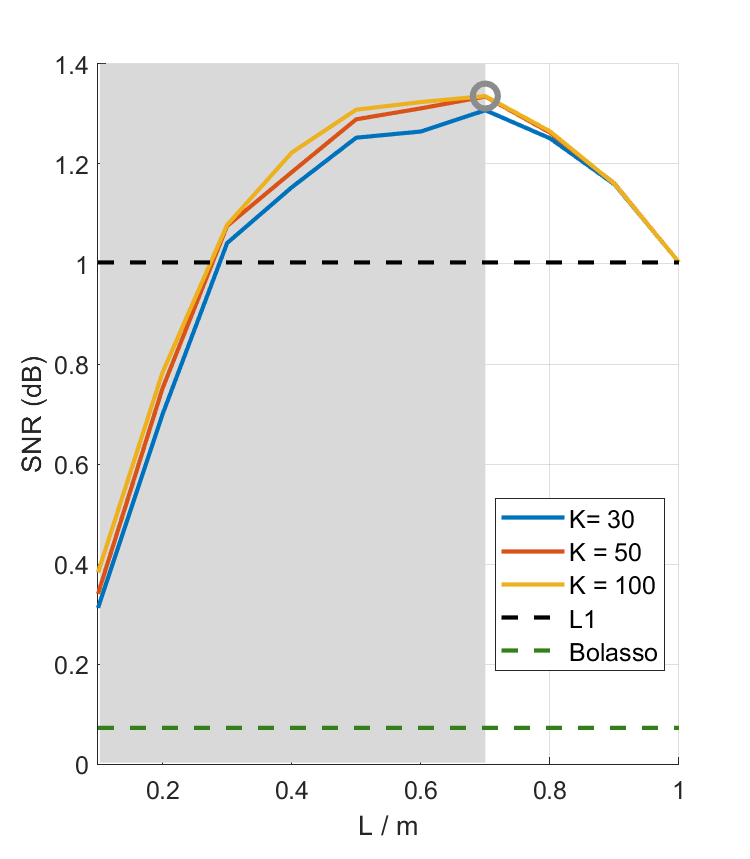}}\\
%
\vspace{-0.5cm}
\subfloat[]
{\raisebox{\dimexpr 3cm-\height}[0pt][0pt]{\rotatebox[origin=c]{90}{$m = 150$}}}
\subfloat[Ours][]
{\label{fg:l11}\includegraphics[trim = {0 0 0 20}, clip, width=0.25\textwidth]{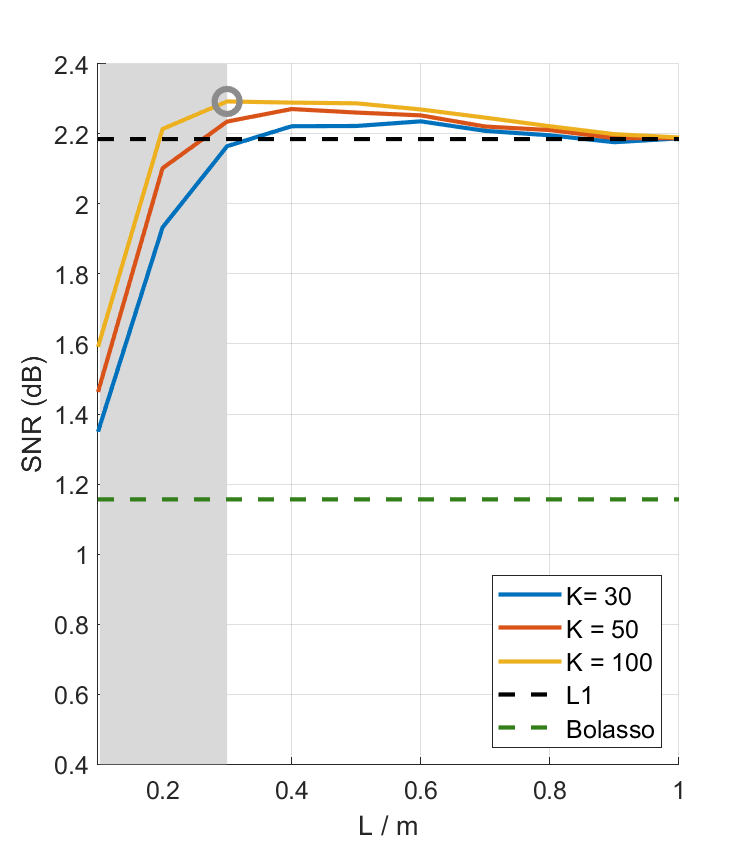}}
\subfloat[Bagging][]
{\label{fg:l12}\includegraphics[trim = {0 0 0 20}, clip, width=0.25\textwidth]{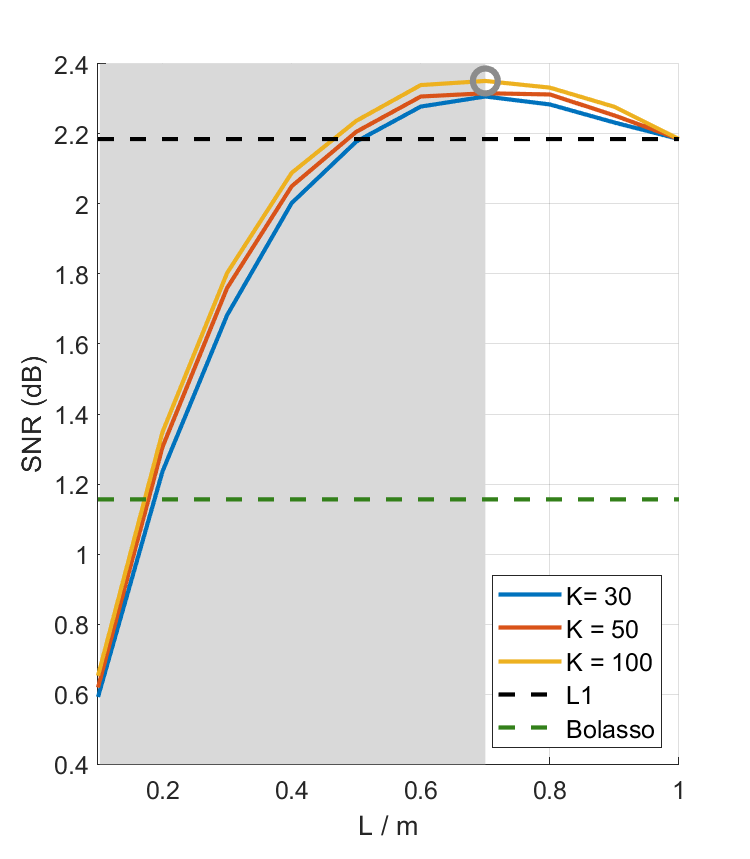}}
\vspace{-0.3cm}
\caption{Performance curves for the 
\textbf{subsampling variations} of JOBS, Bagging and Bolasso with different $L,K$, and $\ell_1$ minimization. $\SNR = 0$ and the number of measurements $m = 50 , 75, 100, 150$ from top to bottom.
The grey circle highlights the peaks of JOBS, Bagging and Bolasso and the grey area highlights the subsampling ratio at the peak point. JOBS requires smaller $L/m$ than Bagging to achieve peak performance. JOBS and Bagging outperform $\ell_1$ minimization and Bolasso when $m$ are small. }
\vspace{-0.3cm}
\label{fg:exp1_var}
\end{figure}
\begin{figure}[!h]
\centering
\parbox{0.27\textwidth}{\centering{JOBS}}
\parbox{0.205\textwidth}{\centering{Bagging}}
\hfil
\vspace{-0.2cm}
\captionsetup[subfigure]{labelformat=empty}
\subfloat[]
{\raisebox{\dimexpr 3cm-\height}[0pt][0pt]{\rotatebox[origin=c]{90}{$m = 200$}}}
\subfloat[Ours][]
{\label{fg:l11}\includegraphics[trim = {0 0 0 20}, clip, width=0.25\textwidth]{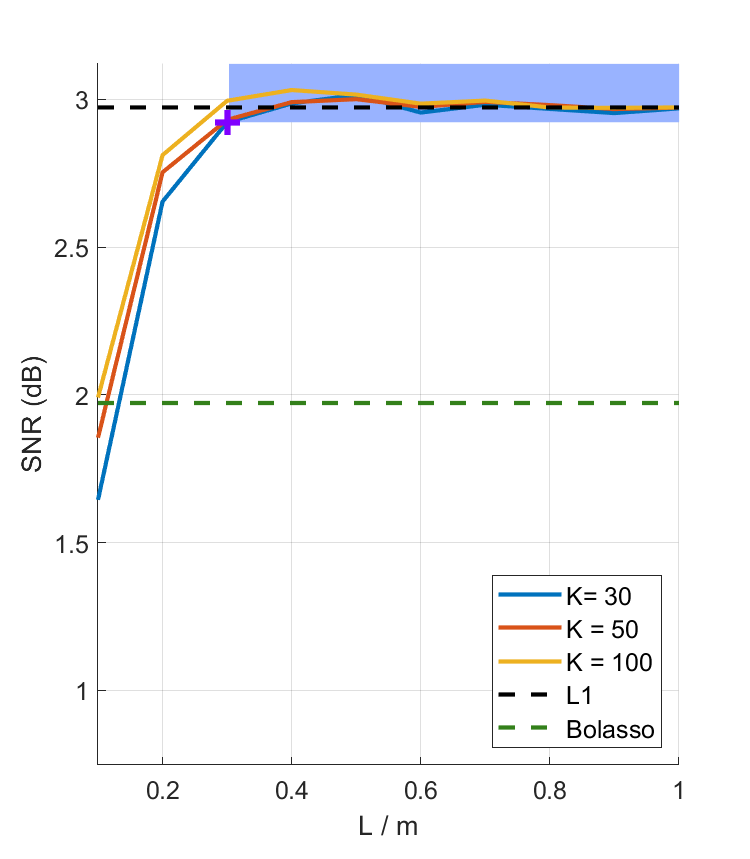}}
\subfloat[Bagging][]
{\label{fg:l12}\includegraphics[trim = {0 0 0 20}, clip, width=0.25\textwidth]{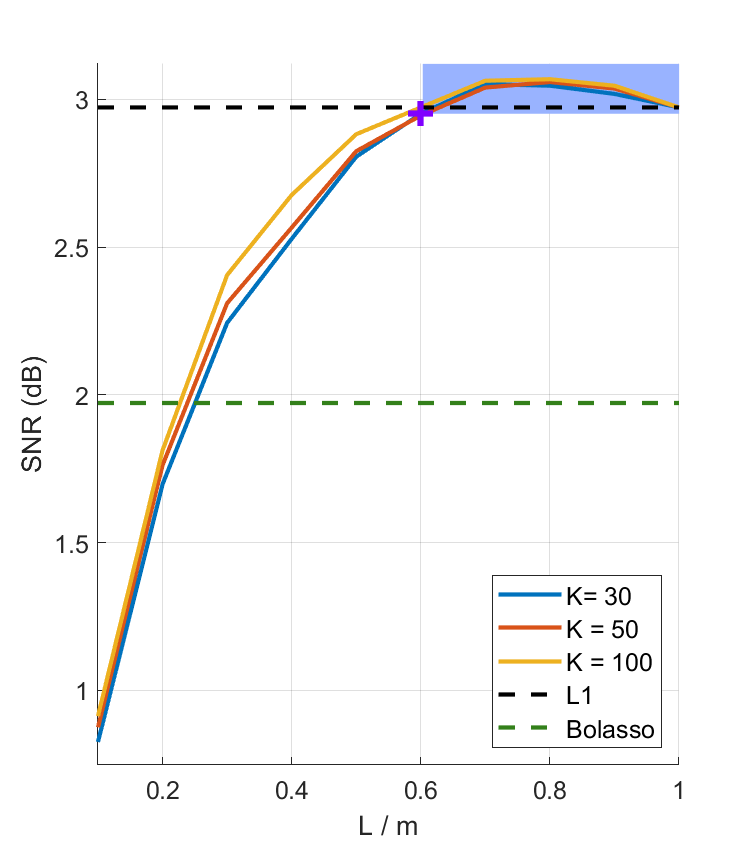}}\\
\vspace{-0.5cm}
\subfloat[]
{\raisebox{\dimexpr 3cm-\height}[0pt][0pt]{\rotatebox[origin=c]{90}{$m = 500$}}}
\subfloat[Ours][]
{\label{fg:l11}\includegraphics[trim = {0 0 0 20}, clip, width=0.25\textwidth]{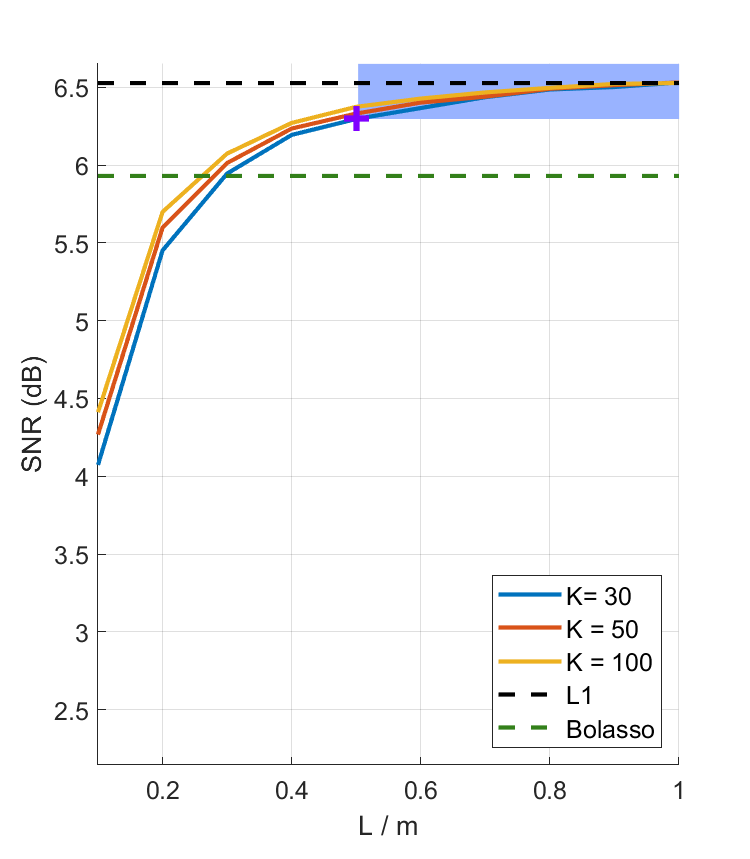}}
\subfloat[Bagging][]
{\label{fg:l12}\includegraphics[trim = {0 0 0 20}, clip, width=0.25\textwidth]{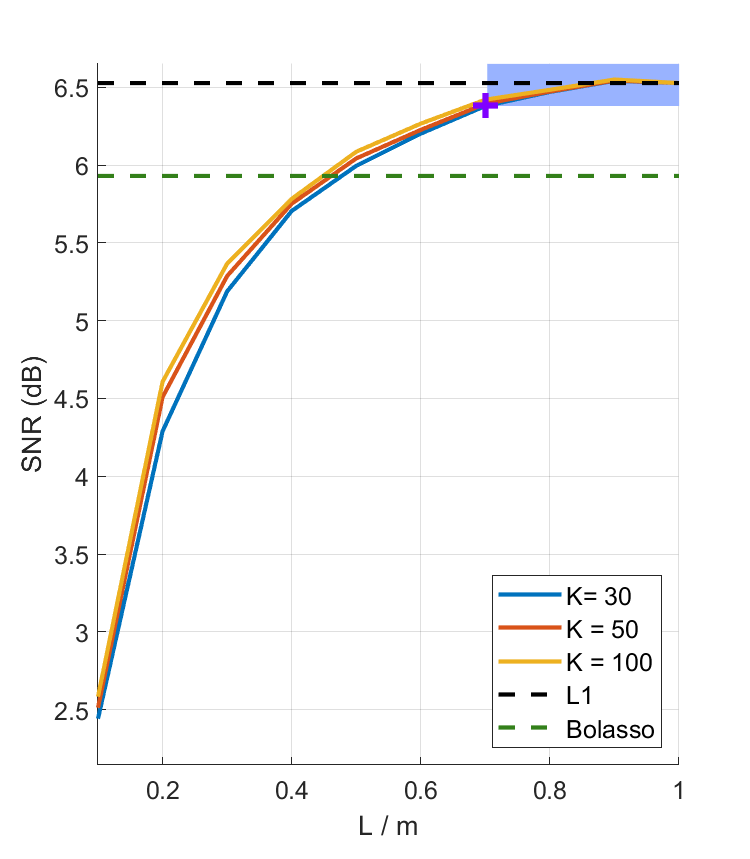}}\\
%
\vspace{-0.5cm}
\subfloat[]
{\raisebox{\dimexpr 3cm-\height}[0pt][0pt]{\rotatebox[origin=c]{90}{$m = 2000$}}}
\subfloat[Ours][]
{\label{fg:l11}\includegraphics[trim = {0 0 0 20}, clip, width=0.25\textwidth]{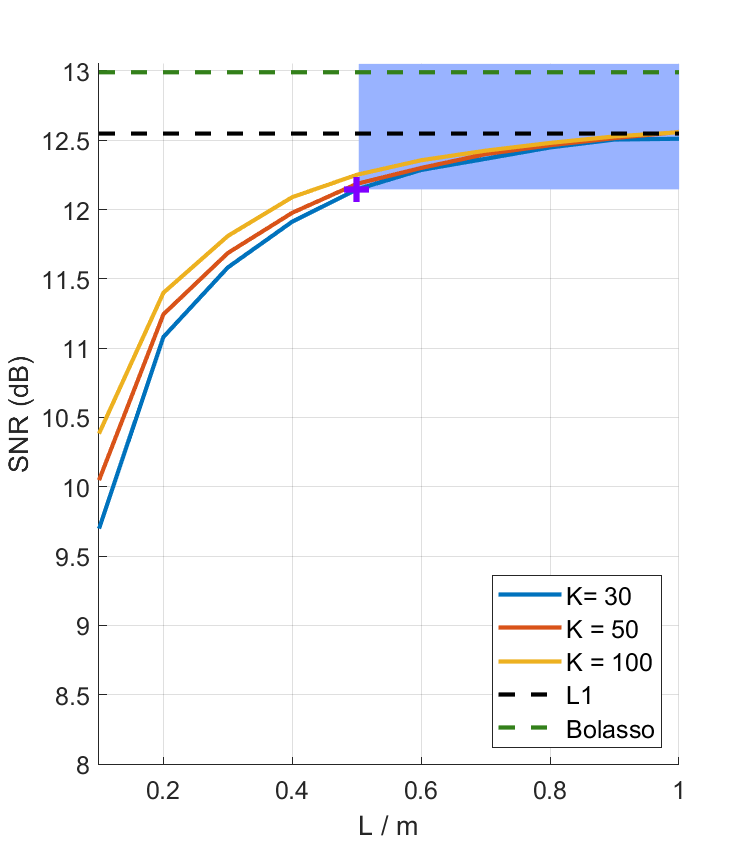}}
\subfloat[Bagging][]
{\label{fg:l12}\includegraphics[trim = {0 0 0 20}, clip, width=0.25\textwidth]{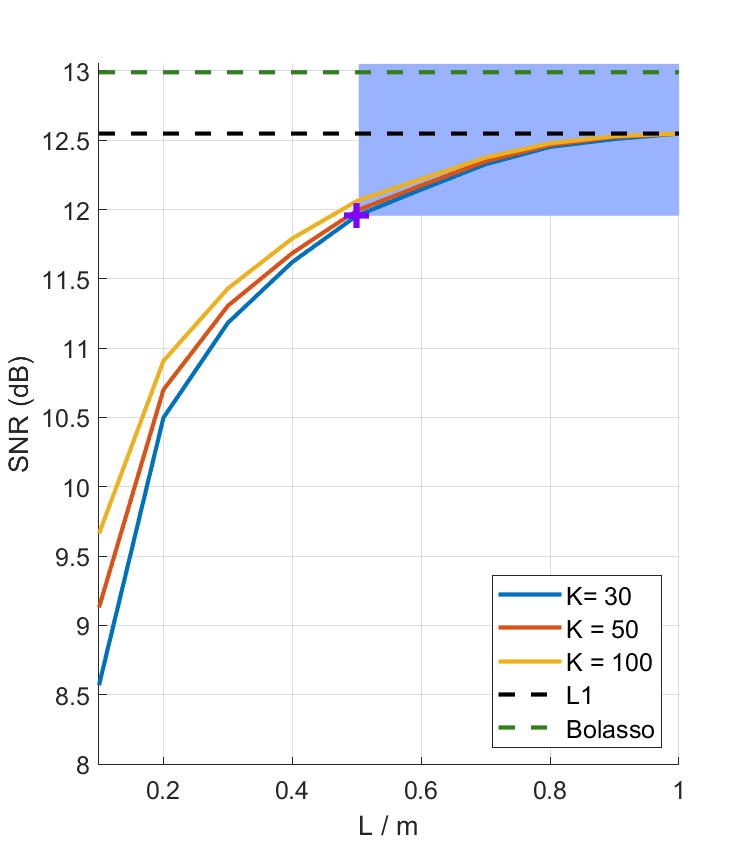}}
\vspace{-0.4cm}
\caption{Performance curves for the 
\textbf{subsampling variations} of JOBS, Bagging and Bolasso with different $L,K$, and $\ell_1$ minimization. $\SNR=0 $ dB and the number of measurements $m = 200 , 500, 2000$ from top to bottom.
The purple highlighted area is where at least $95 \%$ of the best performance achieved and within which the minimum $K \times L $ is illustrated by purple cross.  JOBS requires smaller $L/m$ than Bagging to achieve acceptable performance.
}
\vspace{-0.5cm}
\label{fg:exp1_var_largeM}
\end{figure}

Figure~\ref{fg:exp1_var} depicts the performances of three different algorithms with the same parameters $K,L$ settings. Similar to the case in Figure~\ref{fg:exp1}, we see that both JOBS and Bagging outperforms $\ell_1$ minimization and JOBS achieves the best performance with smaller $L$ than Bagging. Since subsampling potentially contain more information than bootstrap, it also reduces the length of the subsets $L$ for the best performance. For JOBS, the best subsampling ratio $L/m$ at which the peak value is achieved reduces to $20\% - 40 \%$ for small $m$ (ranging from $50 - 150$) , and for Bagging, the optimal subsampling ratio becomes $50\% - 70 \%$.
Figure~\ref{fg:exp1_var_largeM} shows the experiments on large number of measurements ($m = 200, 500, 2000$) with subsampling variation of JOBS, Bagging and Bolasso. The subsampling ratio $L/m$ is $30\%-50\%$ for JOBS and $50\%-70\%$ for Bagging to achieve at least $95\%$ of the best performance, which reaches $\ell_1$ minimization for the subsampling variation.



\begin{figure*}[!h]
\centering
\subfloat[][SNR = 0 dB]
{\label{fg:pv_snr0_var}\includegraphics[trim = {0 0 0 20}, clip, width=0.32\textwidth]
{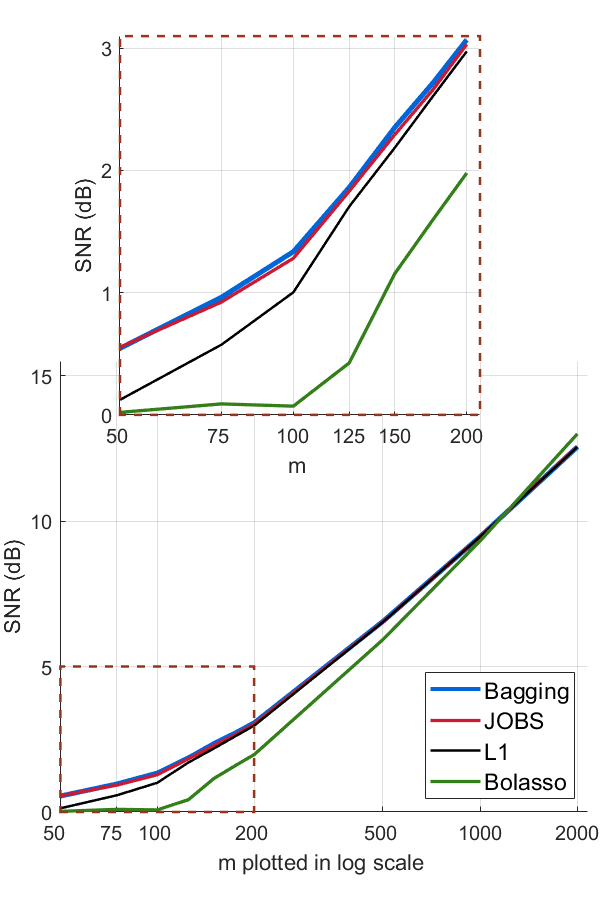}}
\hfill
\subfloat[][SNR = 1 dB]
{\label{fg:pv_snr1_var}\includegraphics[trim = {0 0 0 20}, clip, width=0.32\textwidth]
{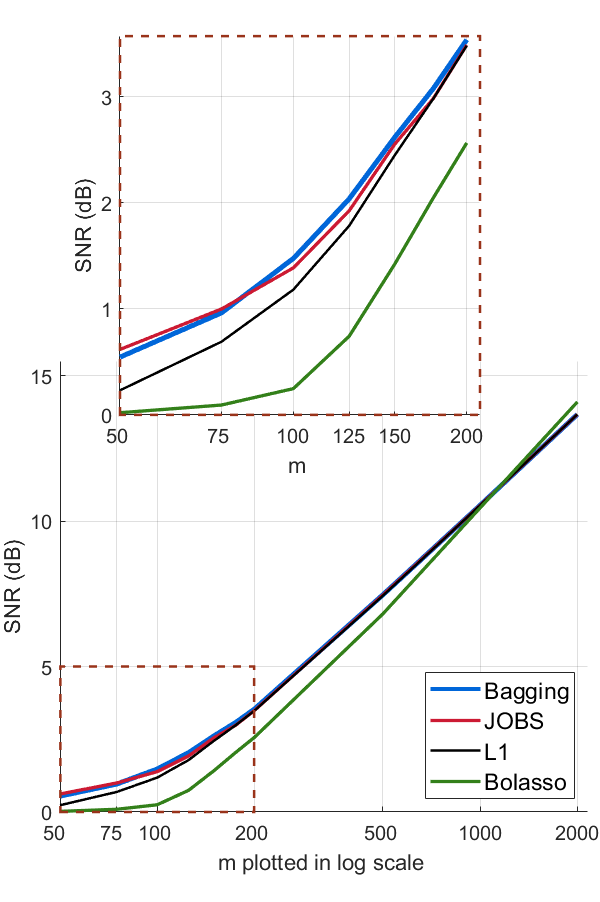}} 
\hfill
\subfloat[][SNR = 2 dB]
{\label{fg:pv_snr2_var}\includegraphics[trim = {0 0 0 20}, clip, width=0.32\textwidth]
{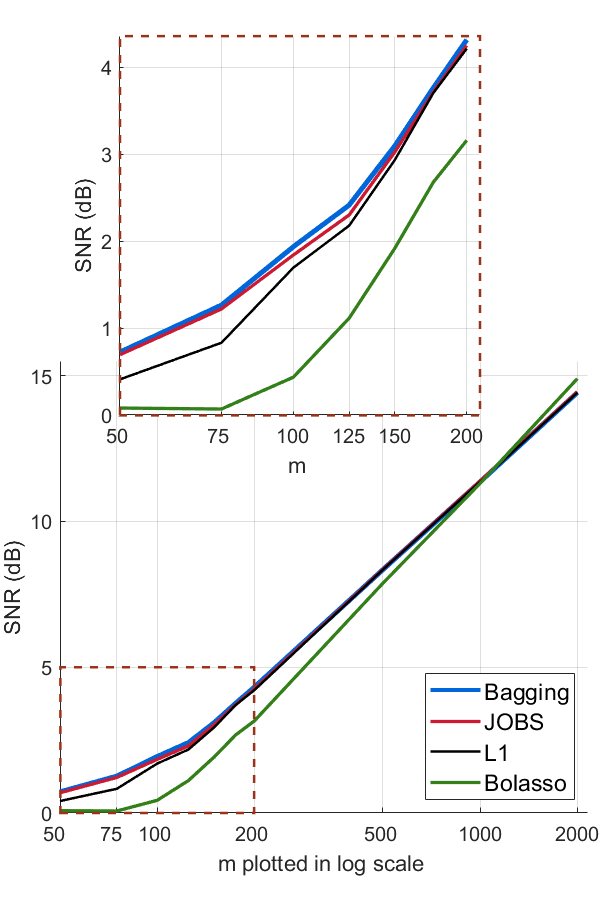}}
\caption{The best performance among all choices of $L$ and $K$ with various number of measurements for \textbf{subsampling variations} of JOBS, Bagging, Bolasso and $\ell_1$ minimization.  While $m$ is small and lower SNR, the margin between JOBS and $\ell_1$ minimization is larger (zoomed-in figures on the top row).}
\label{fg:peaks_snrs_var}
\end{figure*}

Figure~\ref{fg:peaks_snrs_var} 
depicts the best performance for four different recovery scheme: JOBS, Bagging, Bolasso, all in subsampling variations,
and $\ell_1$ minimization with SNR values at $0,1$ and $2$ dB. Similar to Figure~\ref{fig:peaks_wtP},
when the number of measurements $m$ is low, Bagging and our algorithms outperforms $\ell_1$ minimization. The larger the noise level (lower  SNR), the larger the margin. As before, Bolasso only outperforms all other three algorithms when the number of measurements is large. While $L =m$, JOBS and Bagging coincide with the $\ell_1$ minimization.
The optimal values are not that different from the ones in the original bootstrap version in Figure~\ref{fig:peaks_wtP} for the same SNR, especially when $m$ are small. While $m$ is large, the original JOBS and Bagging would need the bootstrap ratio to go above $1$ to achieve the same result as $\ell_1$ minimization and those experiments are not included in this study. We conjecture that the best performance are similar between the original bootstrap scheme and the subsampling variation given the same $m$ with various $K, L$.

With the same $L, K$, the subsampling variation in general gives better performance than bootstrap because more information is likely to be selected. There are two evidences: {\it (i)} While $m$ is small, the optimal subsampling ratios $L/m$ for subsampling variations (in Figure~\ref{fg:exp1_var}) are smaller than the optimal bootstrap ratios (in Figure~\ref{fg:exp1}) for both JOBS and Bagging since the grey and white boundaries are further left in subsampling variations. {\it(ii)} While $m$ is moderate or large, the original JOBS and Bagging start losing advantage to $\ell_1$ minimization whereas for the subsampling variations, JOBS and Bagging both approach to $\ell_1$ minimization with reasonably small $L/m$ and $K$. Good choices of these two parameters are highlighted in the purple regions in Figure~\ref{fg:exp1_var_largeM}.

\section{Conclusion and Future work}
We propose and demonstrate JOBS, which is motivated from a powerful bootstrapping idea and improves robustness of sparse recovery in noisy environments through the usage of a collaborative recovery scheme. We analyze BNSP, BRIP for our methods as well as the sample complexity. We further derive error bounds for JOBS and Bagging.
The simulations results show that our algorithm consistently outperforms Bagging and Bolasso among most choices of parameters $(L,K)$.
JOBS is particularly powerful when the number of measurements $m$ is small. This condition is notoriously difficult, both in terms of improving sparse recovery results and studying the associated theoretical properties. Despite these challenges,
JOBS outperforms $\ell_1$ minimization by a large margin.  JOBS achieves acceptable performance even with very small $L/m$ (around $40\%$ for the original scheme and $30\%$ for the subsampling variation) and relative small $K$ (like $30$ in our experimental study).
The error bounds for JOBS and Bagging show that increasing $K$ will improve the certainty, which is partially validated in the simulation: although it is more computational consuming to choose a larger $K$, increasing $K$ in general gives a better result. Also, for exactly $\sparse-$sparse signals, we have proven that if the RIP condition is satisfied, then for the same $K,L$, JOBS outperforms Bagging. This result matches the large $m$ cases in the simulation, in which the RIP condition should be satisfied.
Future work would include applying the algorithm to dictionary learning and classification.



%

\section{Appendix}
\subsection{BNSP for $\ell_{p,2| \B}$ norm minimization}
Similar to $\ell_{1,2}$ norm in (\ref{eq:lpq}), the definition for $\ell_{p,q}$ norm over block partition $\B$ for vector $\|\x\|_{p,q|\B}$ is defined as:
\begin{equation}
\begin{split}
\|\x\|_{p,q|\B} = & (\sum_{i =1 }^m \| \x \text{\footnotesize{$[\B_i]$}}^{\transpose} \|^p_q )^{1/p} \\
 & = \| ( \| \x \text{\footnotesize{$[\B_1]$}}^\transpose \|_q , ...,\| \x \text{\footnotesize{$[\B_K]$}}^\transpose \|_q ) \|_p
 \end{split}
\end{equation}
In the case when $q=2$, its Block Null Space Property is studied. We recall 
Definition $24$ in~\cite{bnsp} that gives BNSP for the mixed $\ell_{p,2}$ norm in the following theorem. 
\begin{definition}[BNSP for $\ell_{p,2}$ minimization]
\label{p2_block_min}
For any set $\S \subset \{ 1, 2, ..., m\} $ with card$(\S) \leq \sparse$, a matrix $\A$ is said to satisfy the $\ell_p, 0 < p  \leq 1$ block null space property over $\B = \{ d_1, d_2,..., d_m\}$ of order $\sparse$, if
\begin{equation}
\| \v\text{\footnotesize{$[\S]$}}  \|_{p,2|\B} < \| \v \text{\footnotesize{$[\S^c]$}} \|_{p,2|\B},
\end{equation}
for all $\v \in \Null{(\A)} \backslash \{ \zero \}$, where  $\v\text{\footnotesize{$[\S]$}}$ denotes the vector equal to  $\v$ on a block index set $\S$ and zero elsewhere.
\end{definition}
\subsection{Proof of the reverse direction for noiseless recovery}
\label{app:p1ktop1}

\begin{lemma}
\label{lemma:p1ktop1}
 If the MMV problem $\loneminNoNoiseMulti$ , $K > 1$, has a unique solution, it will be of form $\optX = (\optx, \optx, ..., \optx)$, and then there is a unique solution to $\loneminNoNoise$: $\optx$.
\end{lemma}
Let us proof the other direction. If $\loneminNoNoiseMulti$ has a unique solution, the solution must be in the form of $\optX=(\optx, \optx,..., \optx)$, and it implies that $\loneminNoNoise$ has a unique solution $\optx$.

If $\loneminNoNoiseMulti$ has a unique solution, then it is equivalent to say that $\A$ satisfied BNSP of order $\sparse$. For all $\V = (\v_1,\v_2,...,\v_k ) \neq \boldsymbol{O}, \v_\ci \in \Null (\A)$, we have $\forall \ \S, |\S| \leq \sparse, \| \V[\S] \|_{1,2} < \| \V{[\S^c]} \|_{1,2}$. This implies that $\forall \ \V = (\v, \zero, \zero,..., \zero) , \v \in \Null(\A)\backslash \{\zero\}$, BNSP is satisfied. Since in this case, except the first column, all other columns are zero and therefore do not contribute to the group norm. Mathematically, for all $\S$, $\| \V[\S] \|_{1,2} =\| \v \text{\footnotesize{$[\S]$}} \|_1$. We therefore will have the BNSP of order $\sparse$ implies the NSP for $\A$ of order $s$.

\subsection{Proof of Proposition~\ref{prop:factorization}}
\label{pf:prop_factorization}
To proof this proposition, we give alternative form of RIP and BRIP which are stated in the following two propositions.  Alternative form of RIP as a function of matrix induced norm is given as the following:

\begin{proposition}[Alternative form of RIP]
\label{prop_rip}
Matrix $\A$ has $\ell_2$-normalized columns, and $\A \in \R^{m\times n}$, $\S \subset \{1,2,..., n\}$ with size smaller or equal to $\sparse$ and $\A_\S$ takes columns of $\A$ with indices in $\S$.
RIP constant of order $\sparse$ of $\A$, $\delta_{\sparse}(\A)$ is:
\begin{equation}
\label{eq:rip_norm}
\delta_{\sparse} (\A) = \max_{\S \subseteq \{1,2,..., n\} , |\S| \leq \sparse} \| {\A^\transpose_\S } \A_\S - \Identity \|_{2\rightarrow2},
\end{equation}
where $\Identity$ is an identity matrix of size $\sparse \times \sparse$ and $\|\cdot\|_{2\rightarrow2}$ is the induced $2-$norm defined as for any matrix $\A$, $\|\A\|_{2\rightarrow2} = \sup_{\x\neq \zero} \frac{\| \A\x\|_2}{\|\x\|_2}$.
\end{proposition}

The proposition \ref{prop_rip} can be directly derived from the definition of RIP constant. Similarly, we can derive the alternative form of BRIP constant as a function of matrix induced norm.


\begin{proposition}[Alternative form of BRIP]
Let matrix $\A$ have $\ell_2$-normalized columns and let $\B = \{ \B_1, \B_2, ..., \B_n \}$ be the group sparsity pattern that defines the row sparsity pattern, with $\B_i$ contains all indices corresponding to all elements of the $i-$th row. $\T \subseteq \{1, 2,..., n\}$ and $\B(\T) = \{  \B_i, i \in \T\}$ be the subsets that takes several groups with group indices in $\T$. and $\A \in \R^{m\times n}$ with Block-RIP constant of order $\sparse$, $\delta_{ \sparse| \B} (\A) $ is
\begin{equation}
\delta_{ \sparse | \B} = \max_{\T \subseteq  \{1,2,..., n\} , |\T| \leq \sparse } \| \A_{\B(\T) }^\transpose \A_{\B( \T)} - \Identity \|_{2\rightarrow2}.
\end{equation}
\end{proposition}

With loss of generality, let us assume that all columns of $\A$ in the original $\ell_1$ minimization 
have unit $\ell_2$ norms. As a consequence, sub-matrices of $\A$: $\A{[\I_\ci]}$s in general will not guaranteed to have normalized columns. 
 Then, we normalize the columns of these matrices using the following normalization procedure. For $\M \in \R^{ L \times n}$, a matrix that does not have any zero column, $\Q (\M) \in \R^{n \times n} $ is a normalization matrix of $\M$ such that $\M  \Q(\M)$ has $\ell_2$-normalized columns. Clearly, the normalization matrix of $\M$ can be obtained:
 \begin{equation}
 \label{eq:normalization}
 \Q(\M) 
= \diag (\|\m_1 \|^{-1}_2, \|\m_2 \|^{-1}_2,..., \| \m_n \|^{-1}_2  ), 
\end{equation}
where $\m_\ci$ denotes $\ci-$th column of $\M$.

Similary, construct $\Q_{\ci}$s using (\ref{eq:normalization}) to normalize $\A{[\I_\ci]}$s.
 Here, $\A{[\I_\ci]}$s are sub-rows of $\A$, the norm of all columns in $\A{[\I_\ci]}$ are all less than those of $\A$ (which are all equal to $1$). 
Since $\Q_\ci$ contains reciprocals of norm of columns of matrix,   $\Q_\ci$s are diagonal matrices with their diagonals greater or equal to $1$.


Let $\AJ = \text{block}\_\diag( \A{[\I_1]} ,\A{[\I_2]}, ..., \A{[\I_K]})$ and $\QJ = \text{block}\_\diag(\Q_1, \Q_2, ...,\Q_K) $. Then columns of $\AJ \QJ = \text{block}\_\diag(\A{[\I_1]}\Q_1, \A{[\I_2]}\Q_2,..., \A{[\I_K]} \Q_K)$ all have unit norms.
 We consider the BRIP constant for $\AJ $. In this derivation, column selection of a matrix is written as a right multiplication of matrix $\Identity_{\S} (\cdot)$.
\begin{equation*}
\begin{split}
\delta_{ \sparse| \B} & (\AJ)   \\
= &\max_{\T \subseteq  \{1,2,.., n\} , |\T| \leq \sparse}  \| ( \AJ \QJ \Identity_{\B(\T)} )^\transpose  \AJ \QJ \Identity_{\B(\T)} - \Identity \|_{2\rightarrow2} \\
 = & \max_{ \substack{\T \subseteq  \{1,2,.., n\} ,\\ |\T| \leq \sparse}} \ \max_{ \ci } \|  (  \AIci \Q_\ci \Identity_{\T} )^\transpose \   \A{[\I_\ci]} \Q_\ci \Identity_{\T} - \Identity \|_{2\rightarrow2}\\
= &\max_{\substack{\T \subseteq  \{1,2,.., n\} , \\ |\T| \leq \sparse}}   \| \text{block}\_\diag( (\A{[\I_1]} \Q_1 \Identity_{\T})^\transpose \A{[\I_1]} \Q_1 \Identity_{\T} - \Identity,\\
&    \quad \quad ...,   (\A{[\I_K]} \Q_K \Identity_{\T})^\transpose \A{[\I_K]} \Q_K \Identity_{\T} - \Identity) \|_{2\rightarrow2}
\end{split}
\end{equation*}
The induced $2-$norm of a matrix equals to the max singular value of   $\| \D \|_{2\rightarrow2} = \sigma_{\max} (\D)$ and if $\D$ is a block diagonal matrix $\D = \diag(\D_1, \D_2, ..., \D_n) , \sigma_{\max}(\D) = \max \sigma_{\max}(\D_\ci) $. We use this property here. Then
\begin{equation*}
\begin{split}
  \delta_{ \sparse| \B}  &(\AJ) \\
   =&   \  \max_{\substack{\T \subseteq  \{1,2,.., n\} ,\\ |\T| \leq \sparse}}  \max_{ \ci } \|  ( \AIci \Q_\ci  \Identity_{\T} )^\transpose \  \AIci \Q_\ci \Identity_{\T} - \Identity \|_{2\rightarrow2} \\
  =&  \max_{ \ci} \  \max_{\substack{\T \subseteq  \{1,2,.., n\} ,\\ |\T| \leq \sparse}} \|  ( \AIci \Q_\ci  \Identity_{\T} )^\transpose \  \AIci \Q_\ci \Identity_{\T} - \Identity \|_{2\rightarrow2} \\
 =  &  \max_{ \ci} \delta_{\sparse} (\AIci ) .
 \end{split}
 \end{equation*}

\subsection{The Close Form of Birthday Problem}
\label{app:bp}
We generate $L$ samples from $m$ samples uniformly at random with replacement ($L \leq m$). Let $V$ denote the number of distinct samples among $L$ samples. Clearly we have $V \in [1, L]$ and the probability mass function is given by \cite{birthdayIneq}, same as the famous Birthday problem in statistics:
 \begin{equation}
 \label{eq:birthday}
 \begin{split}
  & \P ( V  = v ) = {m \choose v} \sum_{ j  =  0}^ v (-1)^j { v \choose j} (\frac{v -  j}{m})^L , \\
 &  v = 1, 2, ..., L 
 \end{split}
  \end{equation}

In our problem, we are interested in finding the lower bound of $V$ with certainty level $1 - \alpha$ 
\begin{equation}
\label{eq:distinct}
 \P ( V \geq d )   =   \sum_{v = d}^L {m \choose v} \sum_{ j  =  0}^ v (-1)^j { v \choose j} (\frac{v -  j}{m})^L  \geq 1 - \alpha.\\
\end{equation}
Therefore for
\begin{equation}
\label{eq:distinct_alpha}
1 \geq \alpha \geq 
\sum_{v = 0}^{ d-1} {m \choose v} \sum_{ j  =  0}^ v (-1)^j { v \choose j} (\frac{v -  j}{m})^L,
\end{equation}
 equation (\ref{eq:distinct}) is satisfied.


\subsection{Proof of Lemma \ref{lemma:sum_of_rvs}}
\label{sec:proof_lemma_sum_iid}
To prove of this lemma,
we would need 
 Markovs' inequality for non-negative random variables: let $X$ be a non-negative random variable and suppose that $\E X$ exists, for any $t > 0$, we have:
\begin{equation}
\label{eq:markov}
\P\{X > t \} \leq \frac{\E X }{t}.
\end{equation}
We also need the upper bound of the moment generating function (MGF) of random variable $Y$: suppose that $a \leq Y \leq b$, then for all $t \in \R$:
\begin{equation}
\label{eq:ubd_mgf}
\E \exp\{ t Y\} \leq \exp\{ t \E Y + \frac{t^2(b-a)^2}{8}\}.
\end{equation}
Then we start prove the Lemma~\ref{lemma:sum_of_rvs}, for $t > 0$.
\begin{equation*}
\begin{split}
\P & \{ \sum_{i = 1}^n Y_i \geq n \epsilon \} = \P \{ \exp \{ \sum_{i = 1}^n Y_i \}  \geq  \exp \{ n \epsilon \} \}\\
& = \P \{ \exp \{ t \sum_{i = 1}^n Y_i \}  \geq  \exp \{ t n \epsilon \} \};\\
 \end{split}
 \end{equation*}
using the Markov inequality in (\ref{eq:markov}),
 \begin{equation*}
 \begin{split}
 & \leq \exp \{ - t n \epsilon \}  \E \{ \exp \{ t \sum_{i = 1}^n Y_i \} \}  \\
 & = \exp \{ - t n \epsilon \}  \E \{ \Pi_{i = 1}^n \exp \{ t  Y_i \} \} \\
 & = \exp \{ - t n \epsilon \}  \Pi_{i = 1}^n  \E \{ \exp \{ t  Y_i \} \}, \\
 \end{split}
 \end{equation*}
and by upper bound for MGF in (\ref{eq:ubd_mgf})
 \begin{equation*}
 \begin{split}
 & \leq \exp \{ - t n \epsilon \}   ( \exp\{ t \E Y + \frac{t^2(b-a)^2}{8}\})^n\\
  & =  \exp \{ - t n \epsilon +  t n \E Y +  \frac{t^2(b-a)^2n}{8}\}.
\end{split}
\end{equation*}
The right hand side is a convex function respect to $t$, taking the derivative with respect to $t$ and set it zero, we obtain the
optimal $t$, $t^\star =   \frac{ 4 \epsilon - 4 E Y }{( b - a )^2 } $ and the right hand side is minimized: 
\begin{equation*}
 \exp \{ - t^\star n \epsilon +  t^\star n \E Y +  \frac{{t^\star}^2(b-a)^2n}{8}\} = \exp \{ \frac{ -2 n (\epsilon - \E Y )^2}{ ( b - a)^2}  \} .
\end{equation*}


\bibliographystyle{unsrt}
 \bibliography{ieee_jobs}

%
%
%
%
%
\end{document}